\documentclass[journal]{IEEEtran}

\usepackage{color,verbatim}
\usepackage{graphicx}
\usepackage{amsfonts}
\usepackage{amsmath}
\usepackage{mathtools}
\usepackage{amssymb}
\usepackage{bm}
\usepackage{multirow}
\usepackage{accents}
\usepackage{theoremref}
\usepackage{tabularx}
\usepackage{algorithmic}
\usepackage{algorithm}
\usepackage{url}
\usepackage{multicol}
\usepackage{array}
\usepackage[normalem]{ulem}
\usepackage{caption}
\usepackage{subcaption}
\usepackage{xcolor,colortbl}
\usepackage{cite}
\usepackage{amsthm}
\usepackage[hidelinks]{hyperref}

\newtheoremstyle{paperstyle} % Name
  {5pt} % Space above
  {5pt} % Space below
  {} % Body font
  {} % Indent amount
  {\bfseries} % Theorem head font
  {.} % Punctuation after theorem head
  { } % Space after theorem head, ' ', or \newline
  {\thmname{#1}\thmnumber{ #2}\thmnote{ (#3)}} % Theorem head spec (can be left empty, meaning `normal')

\theoremstyle{paperstyle}

\newtheorem{proposition}{Proposition}

\begin{document}
    
    % HEADLINE
    \title{A Trainable Approach to Zero-delay Smoothing Spline Interpolation \thanks{This work was supported by the SFI Offshore Mechatronics grant 237896/O30 and the IKTPLUSS INDURB grant 270730/O70.}}
        \author{
    \IEEEauthorblockN{Emilio Ruiz-Moreno\IEEEauthorrefmark{1}, Luis Miguel L\'opez-Ramos\IEEEauthorrefmark{1},~\IEEEmembership{Member,~IEEE},\\~and Baltasar Beferull-Lozano\IEEEauthorrefmark{1}\IEEEauthorrefmark{2},~\IEEEmembership{Senior Member,~IEEE}\\
    \vspace{5pt}
    \IEEEauthorblockA{\IEEEauthorrefmark{1}WISENET Center, Department of ICT, University of Agder, Grimstad, Norway}\\
    \IEEEauthorblockA{\IEEEauthorrefmark{2} SIGIPRO Department, Simula Metropolitan Center for Digital Engineering, Oslo, Norway}
    }
    \vspace{-10pt}
    }
    \maketitle
    
    % SECTIONS
    % ABSTRACT
\begin{abstract}
The task of reconstructing smooth signals from streamed data in the form of signal samples arises in various applications.
This work addresses such a task subject to a zero-delay response; that is, the smooth signal must be reconstructed sequentially as soon as a data sample is available and without having access to subsequent data.
State-of-the-art approaches solve this problem by interpolating consecutive data samples using splines.
Here, each interpolation step yields a piece that ensures a smooth signal reconstruction while minimizing a cost metric, typically a weighted sum between the squared residual and a derivative-based measure of smoothness.
As a result, a zero-delay interpolation is achieved in exchange for an almost certainly higher cumulative cost as compared to interpolating all data samples together.
This paper presents a novel approach to further reduce this cumulative cost on average. 
First, we formulate a zero-delay smoothing spline interpolation problem from a sequential decision-making perspective, allowing us to model the future impact of each interpolated piece on the average cumulative cost.
Then, an interpolation method is proposed to exploit the temporal dependencies between the streamed data samples.
Our method is assisted by a recurrent neural network and accordingly trained to reduce the accumulated cost on average over a set of example data samples collected from the same signal source generating the signal to be reconstructed.
Finally, we present extensive experimental results for synthetic and real data showing how our approach outperforms the abovementioned state-of-the-art.
\end{abstract}


    % INTRODUCTION 
\section{Introduction} \label{sec:introduction}
%% Intro figure
\begin{figure}
    \centering
    \includegraphics[width=0.98\columnwidth]{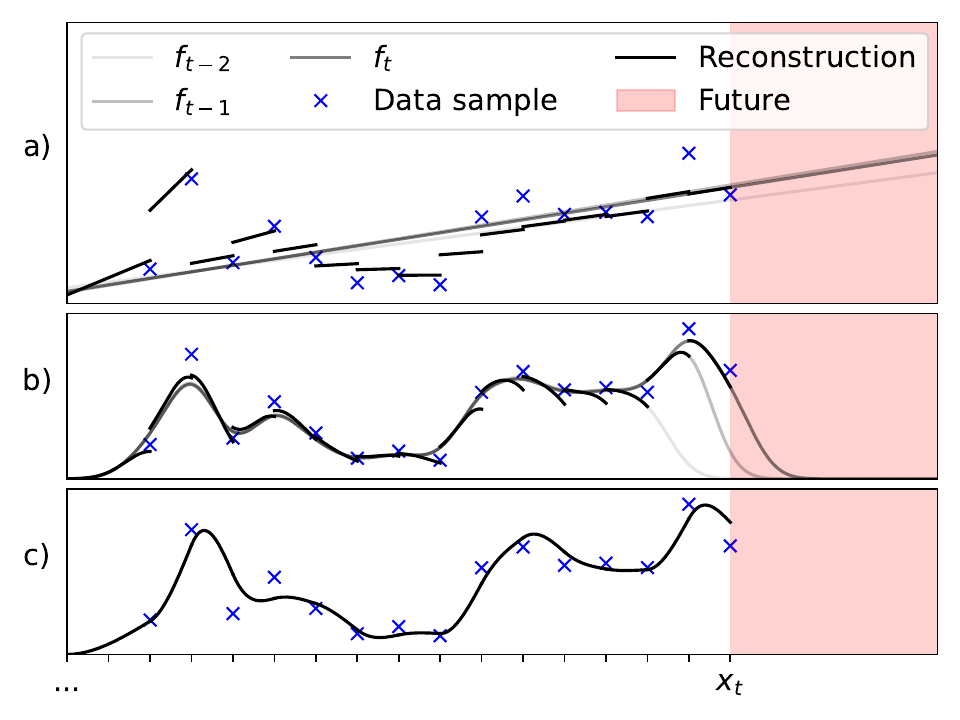}
    \caption{A signal reconstructed from stream data points by different methods and models. The symbols $\textit{x}_t$ and $f_t$ represent the current time and signal estimate, respectively. a) Recursive least squares \cite{plackett1950some,hayes1996statistical} with a linear model. b) NORMA \cite{kivinen2004online} with Gaussian kernels. c) Smoothing myopic interpolation with cubic Hermite splines \cite{de1978practical}.}
    \label{fig:model_comparison}
\end{figure}

Online learning has been studied and applied in a broad range of research fields, including optimization theory \cite{shalev2012online,hazan2016introduction,orabona2019modern}, signal processing \cite{uncini2015fundamentals}, and machine learning \cite{bishop2006pattern,bertsekas2012dynamic,sutton2018reinforcement}.
Within these fields, online methods generating a series of estimates from sequentially streamed data are especially useful to reduce complexity in large-scale problems \cite{bottou2018optimization}, to dynamically adapt to new patterns in the data \cite{mokhtari2016regret}, and to enable acting under real-time requirements \cite{nishihara2017real}.

This work addresses the last one of the previous use cases in the context of signal reconstruction.
Specifically, it investigates the use of online methods with \emph{zero-delay} response for \emph{smooth} signal reconstruction.
First, most physical signals are bounded and smooth due to energy conservation \cite{kosheleva2020physical}; hence it is beneficial to maintain smoothness as a property during signal reconstruction.
Second, the zero-delay requirement demands new portions of the smooth signal to be reconstructed as soon as a new data sample is available. 
Consequently, a reduced constant complexity per iteration \cite{wilf2002algorithms} is required so that the online method is executed at a higher speed than the transmission rate at which the data samples are received.
These requisites are well-motivated since they appear in many practical problems, such as online trajectory planning \cite{bazaz1999minimum,kroger2010line}, real-time control systems \cite{nilsson1998real,gambier2004real}, and high-speed digital to analog conversion \cite{schmidt2017high}, among others.
Although the tasks of estimating smooth signals or delivering a zero-delay response are separately managed by most online methods, handling them together becomes challenging, as we expound next.

Some popular online methods that can be used for smooth signal estimation are online kernel methods \cite{scholkopf2002learning,ruiz2021tracking} and online Gaussian processes \cite{rasmussen2004gaussian,lu2022incremental}.
They aim at yielding a sequence of signal estimates with convergence guarantees or sublinear regret \cite{blum2007learning}.
To this end, they initially propose a signal estimate which is updated (or modified) possibly globally as new data samples arrive.
Their goal is to refine the signal estimate rather than reconstruct new portions of the smooth signal.
Therefore, neither smoothness nor even continuity of the sequentially reconstructed signal is guaranteed.
In fact, any online method not ensuring the smoothness of the reconstruction during the signal estimate update (even when the signal estimate is modeled by smooth functions) suffers from this issue, as illustrated in Fig. \ref{fig:model_comparison}a) and \ref{fig:model_comparison}b).
On the other hand, online interpolation methods can be suitable candidates for the task of smooth signal reconstruction with a zero-delay response.
These methods use piecewise-defined functions to model a sequence of local signal estimates.
Some of these functions allow shaping piecewise-modeled signal estimates that can be updated by assembling a new section (or piece) while guaranteeing the smoothness of the overall sequentially reconstructed signal, as shown in Fig. \ref{fig:model_comparison}c).
Among them, piecewise polynomial functions, also known as \emph{splines}, are arguably the most representative ones \cite{schoenberg1988contributions,schumaker2007spline}.
Actually, splines have been used since ancient times \cite{meijering2002chronology}, long before their mathematical foundations were even established \cite{waring1779vii}, presumably because of their approximation capabilities over functions of arbitrary complexity and ease of use. 

It should be noted that most recursive signal estimation methods modeling function estimates by splines as a basis expansion \cite{wahba1990spline,unser1999splines,unser2005generalized}, suffer from the same issues exposed before.
This is mainly because the smoothness of their signal estimate is directly incorporated into the basis representation and not treated as a set of continuity constraints.
On the contrary, some works \cite{de1986real,dkebski2020real} have explored the task of interpolating sequentially streamed data under real-time requirements by means of splines subject to continuity constraints.
However, the online methods they use involve a multi-step lookahead or shifting window mechanism, which introduces a delay.
Indeed, most online methods for spline interpolation work with local information, for instance, a subset constituted by the last sequentially received data samples. 
In this case, a delayed response allows them to use a larger subset of sequentially received data samples and correct the signal estimates as long as they are updated within the delay limits.
In brief, they can expand the extent of available information at the expense of some delay.
On the other hand, and to the best of our knowledge, the only zero-delay spline interpolation method in the literature is the \emph{myopic} approach, referred to as the ``classical greedy approach'' in \cite{dkebski2020real}, which reduces the delay response to zero by totally ignoring any source of forthcoming information, i.e., a purely local method.
Clearly, there is a research gap on zero-delay spline interpolation methods exploiting additional nonlocal information to achieve a better reconstruction. 
This motivates us to work on the research question of whether it is possible to maintain the zero-delay requirement while efficiently using more information than the myopic approach.

In this paper, we answer affirmatively to the above research question by introducing a novel method for zero-delay smoothing\footnote{Here, the term smoothing refers to a controlled trade-off between fitting the data samples and proposing a smooth signal estimate.} spline interpolation that incorporates a priori information about the dynamics of the signal being reconstructed. 
To this end, we identify the elements of a state space-based sequential decision-making process \cite{frankish2014cambridge} in the context of zero-delay smoothing spline interpolation.
The proposed method relies on a \emph{policy}, i.e., a strategy, that yields a section of the spline (\emph{action}) as a function of the current condition of the so far reconstructed signal and the last received data sample (\emph{state}).
Such a policy consists of a differentiable convex optimization layer (DCOL) \cite{agrawal2019differentiable} on top of a recurrent neural network (RNN) \cite{lecun2015deep,salehinejad2017recent}.
The DCOL allows managing continuity constraints (for any differentiability class) at each interpolation step, thus guaranteeing the smoothness of the signal reconstruction.
The RNN assists the signal estimate update mechanism when appending a new spline section by taking into account the effect of each interpolated section on future interpolation steps.
This aid comes in the form of global data-driven knowledge, and it is tailored to minimize the global cost of the smoothing interpolation problem, on average.
The cost is, in this case, the residual sum of squares plus a weighted derivative-based measure of smoothness.
Lastly, our method is \emph{trainable} in the sense that it uses example time series, i.e., time series sampled from the same signal source generating the signal to be reconstructed, to customize the policy to the temporal dependencies (dynamics) of the signal at hand.

The main contributions of this paper can be summarized as follows: 
\begin{itemize}
    \item We rigorously formulate the problem of smoothing spline interpolation from sequentially streamed data, where each spline section has to be determined as soon as a data sample is available and without having access to subsequent data (zero-delay requirement).
    Due to its nature, it is formulated as a sequential decision-making problem.
    \item As opposed to previously proposed (myopic and not trainable) zero-delay methods, our method trains a policy that aims at minimizing the smoothing interpolation cost metric on average.
    In order to capture the temporal, possibly long-term, dependencies between the streamed data samples and exploit them to reduce further the average cost metric, an RNN able to capture the signal dynamics is incorporated.
    \item The proposed policy guarantees that the reconstructed signal is smooth (a certain number of derivatives are continuous over the interior of the signal domain).  
    This is achieved by adding a DCOL at the output of the RNN and imposing a set of continuity constraints at each interpolation step. 
    In addition, such a layer admits a closed-form evaluation, resulting in improved computational efficiency with respect to off-the-shelf DCOL libraries.
    \item We present extensive experimental results that validate our approach over synthetic and real data.
    Additionally, we show how our approach outperforms the state-of-the-art (namely, myopic) zero-delay methods in terms of the smoothing interpolation average cost metric.
\end{itemize}

The rest of the paper is structured as follows: Sec. \ref{sec:preliminaries} introduces the notation and presents some basic concepts and definitions. 
Then, in Sec. \ref{sec:formulation}, we provide our problem formulation. 
Next, in Sec. \ref{sec:solution} and Sec. \ref{sec:alternative}, we respectively provide a solution, a benchmark, and a baseline. 
Thereafter we experimentally validate our solution in Sec. \ref{sec:experiments}.
Finally, Sec. \ref{sec:conclusion} concludes the paper.

    % PRELIMINARIES
\section{Preliminaries} \label{sec:preliminaries}
In this section, we present the notation and introduce the type of data used in the paper.
Afterward, we address the description of spline-based signal estimates as well as related concepts recurrently appearing in this work.
Finally, we formally describe the smoothing spline interpolation problem, which will be used as a starting point for the formulation of our problem.

%% NOTATION
\subsection{Notation}
Vectors and matrices are denoted by bold lowercase and capital letters, respectively.
Given a vector $\bm{v} = [v_1,\dots,v_C]^\top$, its $c$th component is indicated as $[\bm{v}]_c \triangleq v_c$.
Similarly, given a matrix $\bm{M}\in\mathbb{R}^{R\times C}$, the element in the $r$th row and $c$th column is indicated as $[\bm{M}]_{r,c}$.
The notation $[\bm{v}]_{i:j}$ refers to the sliced vector $[v_i,\dots,v_j]^\top \in \mathbb{R}^{j-i+1}$.
We use Euler's notation for the derivative operator; thus, $D^k_x$ denotes the $k$th derivative over the variable $x$.

%% PROBLEM DATA
\subsection{Problem data} \label{ssec:problem_data}
The data considered in this paper consist of discrete time series, or \emph{series} for short, of $T$ terms each.
We interchangeably refer to the terms of the series as \emph{observations}.
Each \textit{t}th observation $\bm{o}_t\in\mathbb{R}^2$ is described by its time stamp $x_t\in\mathbb{R}$ and its value $y_t\in\mathbb{R}$, i.e., $\bm{o}_t=[x_t,y_t]^\top$.
The observation-associated time stamps are set in strictly monotonically increasing order, i.e., $x_{t-1} < x_t$ for all terms in the series.
Any two consecutive time stamps define a time section $\mathcal{T}_t = (x_{t-1},x_t]$.
Finally, the initial time stamp $x_0$ is set by the user.

%% SPLINE-BASED SIGNAL ESTIMATES
\subsection{Spline-based signal estimates} \label{ssec:splines}
A spline is defined as a piecewise polynomial function.
We denote any spline composed of $T$ piecewise-defined functions, or \emph{function sections}, as
\begin{equation} \label{eq:spline}
    f_{T}(x) = 
    \begin{cases}
        g_1(x), & \text{if } x_0<x\leq x_1 \\
        g_2(x), & \text{if } x_1 < x \leq x_2 \\
        \vdots \\
        g_{T}(x), & \text{if } x_{T-1} < x \leq x_{T}
    \end{cases}
\end{equation}
where every $t$th function section $g_t:\mathcal{T}_t\to\mathbb{R}$ is a linear combination of polynomials of the form
\begin{equation} \label{eq:function_section}
    g_t(x) = \bm{a}^\top_t \bm{p}_t(x) ,
\end{equation}
with combination coefficients $\bm{a}_t \in \mathbb{R}^{d+1}$ and basis vector function $\bm{p}_t:\mathcal{T}_t\to\mathbb{R}^{d+1}$ defined as
\begin{equation} \label{eq:basis_vector}
    \bm{p}_t(x) = \left[ 1,(x-x_{t-1}),\dots,(x-x_{t-1})^d \right]^\top ,
\end{equation}
The integer $d$ denotes the \emph{order} of the spline.
A spline $f_{T}$ is said to have a \emph{degree of smoothness} $\varphi$ if it has $\varphi$ continuous derivatives over the interior of its domain $\text{dom}(f_{T}) = \bigcup_{t=1}^{T}\mathcal{T}_t$.
Next, \textbf{Proposition \ref{prp:continuity}} shows how to enforce continuity up to degree $\varphi \leq d$ in a spline-based signal estimate of order $d$.

\begin{proposition} \label{prp:continuity}
Given a spline expressed as in (\ref{eq:spline}), we can enforce its degree of smoothness to be $\varphi \leq d$ by imposing the following equality constraint
\begin{equation} \label{eq:cc_vf}
    \left[\bm{a}_t\right]_{1:\varphi+1} = \bm{e}_{t-1} ,
\end{equation}
for every $t\in \mathbb{N}^{[1,T]}$, where $\bm{e}_t \in \mathbb{R}^{\varphi+1}$ is a vector such that each of its elements is computed as
\begin{equation} \label{eq:cc_element}
    [\bm{e}_t]_i = \frac{1}{(i-1)!}\sum^{d+1}_{j=1}\left[\bm{a}_{t}\right]_j
    \, u_{t}^{j-i}
    \prod^{i-1}_{k=1}(j-k) ,
\end{equation}
with $u_t \triangleq x_t - x_{t-1}$, and with the exception of $\bm{e}_0$, which determines the initial conditions of the reconstruction and can be either calculated or set by the user. 
\newline
\indent\textit{Proof}: see Appendix \ref{apx:continuity}.
\end{proposition}

%% SMOOTHING SPLINE INTERPOLATION
\subsection{Smoothing spline interpolation} \label{ssec:smoothing_spline_interpolation}
Consider the space $\mathcal{W}_\rho$ of functions defined over the domain $(x_0,x_{T}] \subseteq \mathbb{R}$ with $\rho-1$ absolutely continuous derivatives and with the $\rho$th derivative square integrable. 
Then, given a whole series of observations $\{\bm{o}_t\}^{T}_{t=1}$ with $T\geq\rho$ and a positive hyperparameter $\eta$, we can formulate the following batch optimization problem 
\begin{equation} \label{eq:smoothing_interpolation}
    \underset{f\in \mathcal{W}_\rho}{\text{ min }} \sum^{T}_{t=1} \left( f(x_t) - y_t \right)^2 + \eta \int^{x_{T}}_{x_0} \left( D^\rho_x f(x) \right)^2 \, dx
\end{equation}
known as smoothing spline interpolation \cite{reinsch1967smoothing,wood2006generalized}.
The name is due to the unique solution to the optimization problem \eqref{eq:smoothing_interpolation} being a spline conformed of $T$ function sections, as in \eqref{eq:spline}.
More specifically, the solution of \eqref{eq:smoothing_interpolation} is a spline of order $2\rho-1$ with $2\rho-2$ continuous derivatives and natural boundary conditions \cite{wahba1990spline}.
The hyperparameters $\eta$ and $\rho$ control the smoothness of such a solution. 
Particularly, the integer $\rho$ dictates the minimum required degree of smoothness of the search function space $\mathcal{W}_\rho$ and the type of regularization\footnote{Our experimental setup focuses on $\rho=2$, a common choice in practice, which penalizes excessive curvature in the spline. Applications with $\rho>2$ can also be found, e.g., trajectory planning tasks \cite{fan2015realtime}. However, they are out of the scope of this paper, as we justify in the ensuing Sec. \ref{ssec:policy_configuration}.} (second term in \eqref{eq:smoothing_interpolation}).  
Regarding $\eta$, it controls the trade-off between the squared sum of vertical deviations of the signal estimate from the data and the regularization term.
Notice that as $\eta\to0$, the solution of \eqref{eq:smoothing_interpolation} approaches the interpolation spline while as $\eta\to\infty$, it tends to the polynomial of order $\rho-1$ that best fits the observations in the least-squares sense.

On the other hand, note that the structure of the solution of the problem (\ref{eq:smoothing_interpolation}), being a natural spline, arises organically rather than being imposed in advance.
This is a direct consequence of its batch formulation allowing us to delimit the search space $\mathcal{W}_\rho$ to splines of order $d$ and degree of smoothness $\varphi$ satisfying $2\rho-1 \leq d$ and $\rho-1 \leq \varphi \leq 2\rho-2$ without loss of optimality.
From a practical perspective, it is sufficient to choose the minimum required order and degree of smoothness, thus reducing the model's complexity.
However, this trait is not necessarily present in online settings.
That is, the smoothness of the solution does not arise naturally using online methods, and it has to be enforced.
So here, the choice of the spline order and degree of smoothness is rather user-defined or task-oriented.

    % PROBLEM FORMULATION
\section{Problem formulation} \label{sec:formulation}
Once the problem data, the description of spline-based signal estimates, and the smoothing spline interpolation problem have been introduced in Sec. \ref{sec:preliminaries}, we are ready to formalize the main task of this paper, namely the trainable zero-delay smoothing spline interpolation problem.
This section fully describes the aforementioned task from a data-driven sequential decision-making perspective by introducing a suitable dynamic programming (DP) \cite{bellman1966dynamic} framework.
To this end, we first model the \emph{environment}, define the state space and action space, and delimit a suitable family of candidate policies.
Then we introduce the \emph{total cost} and formulate the above task as the problem of finding the policy incurring the lowest total cost on average.

%% CHARACTERIZATION OF THE PROBLEM DATA
\subsection{Characterization of the problem data}
In our problem, the data described in Sec. \ref{ssec:problem_data} are observed sequentially.
Before every \textit{t}th time step, the observation about to be received $\bm{o}_t$ remains undetermined but still governed by the dynamics of the environment.
In this work, we model the dynamics of the environment as a random process $Y(\omega,x)$, where $\omega$ is a sample point from a sample space $\Omega$, and $x$ is a value within an index set $\mathcal{X} \subseteq \mathbb{R}$, in this case, time.
At time $x_t$, all possible outcomes form a random variable $Y(\omega,x_t)$ or $y_t$ for short.
If the $m$th sample is considered at time $x_t$, the outcome has a value denoted by $Y(\omega_m,x_t)$ or simply $y_{m,t}$. 
Consequently, if a discrete set of time stamps is chosen, i.e., $\mathcal{X}=\{x_1,\dots,x_{T}\}$, $T$ random variables can be formed, and all the information about the discrete random process $Y_\mathcal{X}$ is contained in the joint probability density function $P_{Y_\mathcal{X}}$.

%% STATE SPACE
\subsection{State space}
At every time step $t$, we encode a snapshot of the observable environment and the condition of the so-far reconstructed signal in a vector-valued variable called state.
With $\mathcal{S}$ denoting the state space, each \textit{t}th state $\bm{s}_t\in\mathcal{S}$ is constituted by the corresponding observation $\bm{o}_t$, and the condition at which the reconstruction was left, which is specified by the vector $\bm{e}_{t-1}$ whose components are given as in \eqref{eq:cc_element}, and the time instant $x_{t-1}$.
Formally, every \textit{t}th state $\bm{s}_t$ is expressed as $\bm{s}_t = [ \bm{o}_t^\top, \bm{e}_{t-1}^\top,x_{t-1} ]^\top$.
Since every state $\bm{s}_t$ is uniquely determined once the spline coefficients $\bm{a}_{t-1}$ are fixed, we can explicitly describe the state update mechanism, by means of a deterministic mapping, as
\begin{equation} \label{eq:state_update_mechanism}
    \bm{s}_{t+1} = F\left(\bm{s}_t,\bm{a}_t,\bm{o}_{t+1}\right) .
\end{equation}
Formalizing the state update mechanism in \eqref{eq:state_update_mechanism} allows us to identify all visitable states seamlessly.

%% ACTION SPACE
\subsection{Action space} \label{ssec:action_space}
Immediately after receiving the \textit{t}th observation, we propose a function section as in (\ref{eq:function_section}), and we implicitly select the spline coefficients $\bm{a}_t$.
This is because the function section is determined as soon as $\bm{a}_t$ is chosen (the basis vector defined in \eqref{eq:basis_vector} is given).
From this point of view, selecting the spline coefficients of a function section can be understood as an action.
Any valid action generates a function section of the same order $d$ as the spline reconstruction. 
Formally, $\bm{a}_t \in \mathcal{A} \subseteq \mathbb{R}^{d+1}$ for all the \textit{t}th terms, where $\mathcal{A}$ denotes the action space.
However, if we want a reconstructed spline that is continuous up to the $\varphi$th derivative, not all valid actions are appropriate. 
In our context, for any \textit{t}th action to be deemed \emph{admissible} (or feasible), it must satisfy the constraint in (\ref{eq:cc_vf}).
Notice that the set of admissible actions depends on the current state.
Therefore, we accordingly denote the admissible action space as $\mathcal{A}(\bm{s}_t)$.

%% POLICY SPACE
\subsection{Policy space} \label{ssec:policy_space}
A policy $\pi=\left\{\bm{\mu}_t:\mathcal{S}\to\mathcal{A}\right\}_{t\in\{1,2,\dots\}}$ consists of a sequence of functions that map states into actions.
Policies are more general than actions because they incorporate the knowledge of the state.
However, notice that not all policies return admissible actions.
Only the policies that satisfy $\pi(\bm{s}_t) \in \mathcal{A}(\bm{s}_t)$ for all time steps are termed \emph{admissible policies}.
Separately, \emph{stationary policies} are policies that do not change over time, i.e., $\bm{\mu}\equiv\bm{\mu}_t = \bm{\mu}_{t+1}$ for all time steps.
Hence, a stationary policy is unequivocally defined by the mapping $\bm{\mu}$.
Stationary policies are suitable for making decisions in problems with a varying horizon (varying number of time steps), assuming usually stationary environments. 

These arguments motivate the use of admissible stationary policies.
However, the space of admissible stationary policies is huge, and therefore, the problem of finding the most adequate policy within it can be overwhelmingly complex.
Policy approximation techniques help reduce the pool of candidate policies by restricting them to a certain family of policies.
These techniques tend to work best (in the sense of providing an adequate policy) when the problem has a clear structure that can be accommodated into the policy.
In our case, we aim to incorporate the temporal dependencies across the observations into the policy, as well as the notion of smoothness discussed in Sec. \ref{ssec:smoothing_spline_interpolation}. 
To this end, we resort to parametric policy approximation \cite{bertsekas2012dynamic} denoting any approximated stationary policy as $\bm{\mu}_{\bm{\theta}}$, where the vector $\bm{\theta}\in\mathbb{R}^P$ contains the $P$ parameters constituting the aforenamed policy.
The set $\Pi$ of parametric stationary policies that return admissible actions is, therefore, the space of policies of interest to this work.

%% TOTAL EXPECTED COST
\subsection{Total expected cost} \label{ssec:total_expected_cost}
The following \textbf{Proposition \ref{prp:additivity}} shows that the smoothing spline interpolation objective introduced in Sec. \ref{ssec:smoothing_spline_interpolation}, equation \eqref{eq:smoothing_interpolation}, can be expressed as a summation where each term depends on a single action, resembling the sequence of instantaneous costs in a typical DP formulation.

\begin{proposition} \label{prp:additivity}
The objective of the optimization problem \eqref{eq:smoothing_interpolation} can be equivalently computed additively as 
\begin{equation} \label{eq:additivity}
    \sum^{T}_{t=1} \left(\bm{a}^\top_t\bm{p}_t(x_t)-y_t\right)^2 + \eta\,\bm{a}^\top_t\bm{M}_t\bm{a}_t ,
\end{equation}
where $\bm{M}_t\in\textbf{S}^{d+1}_+$ with elements given by
\begin{equation} \label{eq:definition_M}
\left[\bm{M}_t\right]_{i,j} =
\begin{cases}
    0 & \text{if } i\leq \rho \text{ or } j\leq \rho \\
    \frac{u_t^{i+j-2\rho-1}}{i+j-2\rho -1} \prod^\rho_{k=1}(i-k)(j-k) & \text{otherwise},
\end{cases}
\end{equation}
being $u_t\triangleq x_t - x_{t-1}$. 
\newline
\indent\textit{Proof}: See Appendix \ref{apx:additivity}.
\end{proposition}

Based on \textbf{Proposition \ref{prp:additivity}}, we can express the objective of the smoothing spline interpolation problem \eqref{eq:smoothing_interpolation}, as the total cost $\sum^{T}_{t=1} \kappa(\bm{s}_t,\bm{a}_t)$, with \emph{cost} $\kappa:\mathcal{S}\times\mathcal{A}\to\mathbb{R}$ given by
\begin{equation} \label{eq:cost}
    \kappa(\bm{s}_t,\bm{a}_t) = \left(\bm{a}^\top_t\bm{p}_t(x_t)-y_t\right)^2 + \eta\,\bm{a}^\top_t\bm{M}_t\bm{a}_t ,
\end{equation}
where $\bm{M}_t$ is constructed as in (\ref{eq:definition_M}).
This is because each \textit{t}th state-action pair contains all necessary information.
From here and under a given policy of interest $\bm{\mu}_{\bm{\theta}}$, as described in Sec. \ref{ssec:policy_space}, the metric 
\begin{equation} \label{eq:total_expected_cost}
    \underset{y_t \sim P_{Y_\mathcal{X}}}{\mathbb{E}} \left[ \sum^{T}_{t=1} \kappa \left( \bm{s}_t, \bm{\mu}_{\bm{\theta}}(\bm{s}_t) \right) \right] ,
\end{equation}
denotes the \emph{total expected cost} incurred by following such a policy from a given initial state $\bm{s}_0$, and traveling all the remaining states $\bm{s}_t \in \mathcal{S}_t$ via \eqref{eq:state_update_mechanism}.
The expectation in (\ref{eq:total_expected_cost}) is performed over the random process modeling the dynamics of the environment through the observations within the states.

%% POLICY SEARCH BY COST OPTIMIZATION
\subsection{Policy search by cost optimization} \label{ssec:policy_search}
Computing the expectation in \eqref{eq:total_expected_cost} is computationally expensive or even intractable when the underlying random process generating the series of observations is unknown.
Instead, we can rely on sample average approximation of example series collected from past realizations of the process. 
The sample average approaches the expectation as the number of examples grows.
In this way, we can determine a data-driven policy by solving the following optimization problem
\begin{subequations}  \label{eq:policy_search}
\begin{align}
    \text{arg}\underset{\bm{\mu}_{\bm{\theta}}\in\Pi}{\text{ min }}& \sum_{m=1}^M \sum^{T}_{t=1} \kappa \left( \bm{s}_{m,t},\bm{\mu}_{\bm{\theta}}(\bm{s}_{m,t}) \right) \label{seq:policy_search_objective} \\
    \text{s. to: }& \bm{s}_{m,t} = F\left(\bm{s}_{m,t-1},\bm{\mu}_{\bm{\theta}}(\bm{s}_{m,t-1}),\bm{o}_{m,t}\right), \forall m,t , \\
    & \bm{\mu}_{\bm{\theta}}(\bm{s}_{m,t}) \in \mathcal{A}(\bm{s}_{m,t}) , \forall m,t ,
\end{align}
\end{subequations}
where the integer $M$ denotes the number of example series, indexed by $m$, and where all the initial states $\bm{s}_{m,0}$ as well as all observations $\bm{o}_{m,t}$ are given.
    % PROPOSED SOLUTION
\section{Proposed solution} \label{sec:solution}
The previous Sec. \ref{sec:formulation} has provided the necessary definitions and considerations to arrive at a rigorous problem formulation. 
An exact solution to the problem \eqref{eq:policy_search} is probably impossible to obtain in practice, mainly due to the complexity of the search space $\Pi$.
There are multiple possibilities regarding the policy approximation and optimization techniques that can be taken towards obtaining a near-optimal solution to \eqref{eq:policy_search}.
This section presents a specific set of design choices based on the current state-of-the-art.
In particular, we rely on a policy parametrization through cost parametrization technique, borrowed from the DP literature, in synergy with an RNN architecture. 
Then we make use of backpropagation through time (BPTT) \cite{werbos1990backpropagation}, a gradient computation technique borrowed from the deep learning literature \cite{goodfellow2016deep}. 
Our proposed solution can effectively solve the problem formulated in Sec. \ref{ssec:policy_search} for $\rho\leq2$.
The remaining configurations manifest instability issues, and even though they may be solvable, they lie outside of the scope of the current paper as further discussed in the following Sec. \ref{ssec:policy_configuration}.

Future developments in the DP or deep learning areas, such as new policy approximation approaches, neural architectures, or optimizers, can possibly render the techniques proposed in this section obsolete but will not affect the validity of the problem formulated in Sec. \ref{ssec:policy_search}. 

% POLICY FORM
\subsection{Policy form} \label{ssec:policy_form}
\begin{figure}
    \centering
    \includegraphics[width=0.9\columnwidth]{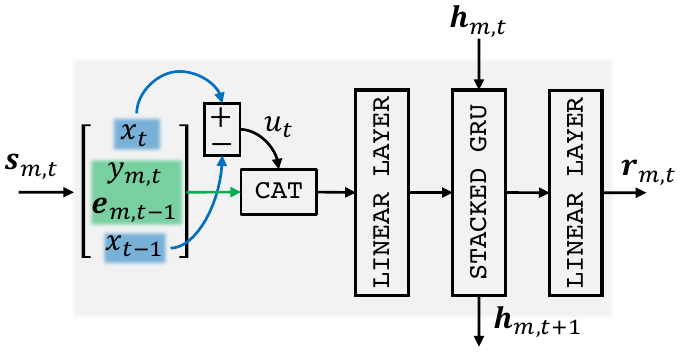}
    \caption{Visual representation of our RNN architecture satisfying the relation in \eqref{eq:rnn}. The elements onto the gray shaded area constitute the mapping $R_{\bm{\theta}'}$. The CAT cell stands for a concatenation operation.}
    \label{fig:rnn_architecture}
\end{figure}
Parametric policy approximation via parametric cost function approximation (CFA) \cite{bertsekas2012dynamic} is a method that seeks through the policy space, in our case $\Pi$, among those policies defined as an optimization problem with parametrized objectives.
In this work, we are interested in CFA-based policies of the form 
\begin{equation} \label{eq:parametric_policy}
    \bm{\mu}_{\bm{\theta}}(\bm{s}_{m,t}) = \text{arg}\underset{\bm{a}\in\mathcal{A}(\bm{s}_{m,t})}{\text{min}} \left\{ \kappa (\bm{s}_{m,t},\bm{a}) + J_{\bm{\theta}}(\bm{s}_{m,t},\bm{a};\bm{h}_{m,t}) \right\} ,
\end{equation}
where the map $\kappa$ denotes the cost described in \eqref{eq:cost}, and the mapping $J_{\bm{\theta}}:\mathcal{S}\times\mathcal{A}\times\mathbb{R}^H\to\mathbb{R}$ is a parametric \emph{cost-to-go} approximation involving $P$ parameters contained in the vector $\bm{\theta}$.
Regarding the vector $\bm{h}_{m,t} \in \mathbb{R}^H$, it represents a \emph{latent state} value at the \textit{t}th time step of an \textit{m}th example series.
The latent state may encode relevant information from past observations and can be viewed as a policy memory \cite{peshkin2001learning,schafer2008reinforcement,zhang2016learning}.

We aim for a cost-to-go approximation $J_{\bm{\theta}}$, which penalizes those actions that are distant from the output of a certain RNN.
The main reason behind this approach is that an RNN that successfully captures the temporal dynamics of the environment has the potential to pull towards actions that yield a low expected total cost. 
So, it is constructed as follows
\begin{equation} \label{eq:cost-to-go_approximation}
    J_{\bm{\theta}}(\bm{s}_{m,t},\bm{a};\bm{h}_{m,t}) = \lambda \left\Vert \bm{a} -
    \begin{bmatrix}
        \bm{0}_{\varphi + 1} \\
        \bm{r}_{m,t}
    \end{bmatrix}
    \right\Vert^2_2 ,
\end{equation}
where $\lambda\in\mathbb{R}_+$.
The vectors $\bm{r}_{m,t} \in \mathbb{R}^{d-\varphi}$, and $\bm{h}_{m,t} \in \mathbb{R}^H$ represent the outputs and latent state of an RNN, $R_{\bm{\theta}'} : \mathcal{S}_t \times \mathbb{R}^H \to \mathbb{R}^{d-\varphi} \times \mathbb{R}^H $, respectively. 
They are obtained from the following relation
\begin{equation} \label{eq:rnn}
    R_{\bm{\theta}'}(\bm{s}_{m,t};\bm{h}_{m,t}) = 
    \begin{bmatrix}
        \bm{r}_{m,t} \\
        \bm{h}_{m,t+1}
    \end{bmatrix} ,
\end{equation}
with $\bm{\theta} = [\lambda,{\bm{\theta}'}^\top]^\top$ and $\bm{\theta}' \in \mathbb{R}^{P-1}$, exemplified in Fig. \ref{fig:rnn_architecture}.
From now on, we refer to the policy in \eqref{eq:parametric_policy} with cost-to-go as in \eqref{eq:cost-to-go_approximation} as the RNN-based policy.
Finally, notice that besides being parametric, the RNN-based policy is admissible and stationary by design.

% POLICY EVALUATION
\subsection{Policy evaluation} \label{ssec:policy_evaluation}
Evaluating the proposed policy involves solving the optimization problem stated in \eqref{eq:parametric_policy}.
Notice that both the cost $\kappa$ built as in \eqref{eq:cost}, and the cost-to-go approximation $J_{\bm{\theta}}$ described in \eqref{eq:cost-to-go_approximation}, are convex with respect to the actions and hence, the objective in \eqref{eq:parametric_policy} is convex too.
Moreover, the admissible action set, described in Sec. \ref{ssec:action_space}, is convex.
Therefore, the optimization problem in \eqref{eq:parametric_policy} is convex thus, any locally optimal action is globally optimal \cite{boyd2004convex}.

Additionally, the optimization problem in \eqref{eq:parametric_policy} has been designed to admit a closed-form solution.
Closed-form evaluations can usually be computed faster and more precisely than solutions obtained from numerical methods, and thus, they are more suitable under zero-delay requirements.
See Appendix \ref{apx:closed-form_policy_evaluation} for the derivation of the closed-form evaluation.

% POLICY TRAINING
\subsection{Policy training}
\begin{figure}
    \centering
    \includegraphics[width=0.8\columnwidth]{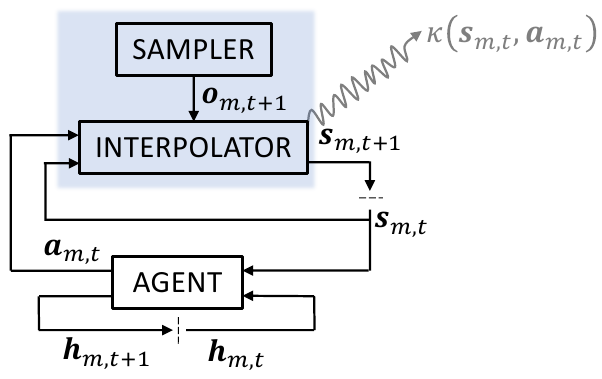}
    \caption{Rolled representation of the environment-agent system. The sampler cell samples the random process modeling the dynamics of the environment. The interpolator cell performs the reconstruction, evaluates the cost, and updates the states. The agent cell contains the policy and the RNN. The blue-shaded area encompasses the environment. The cost (in gray) is used during the training phase but not for evaluation.}
    \label{fig:dp_scheme}
\end{figure}

As explained in Sec. \ref{ssec:policy_form}, we have reduced the search space of problem \eqref{eq:policy_search} by restricting the policy space to a family of policies of the form given in \eqref{eq:parametric_policy}.
Specifically, from searching a function $\bm{\mu}_{\bm{\theta}}$ in the function space $\Pi$, we have narrowed the problem down to that of finding a vector $\bm{\theta}$ in the vector space $\mathbb{R}^P$.
In fact, tuning the proposed policy parameters by solving the optimization problem \eqref{eq:policy_search} is commonly referred to as policy training.
Unfortunately, the objective (\ref{seq:policy_search_objective}) is non-convex with respect to the parameters in $\bm{\theta}$.
As a reasonable solution, we rely on a gradient-based optimizer aiming to converge to a high-performance local minimum.

From a deep learning perspective, the policy evaluation presented in Sec. \ref{ssec:policy_evaluation} can be understood as a forward pass of a DCOL on top of an RNN, and hence, it is trained using BPTT via automatic differentiation \cite{paszke2017automatic}.
This point of view is schematized in Fig. \ref{fig:dp_scheme}, where traveling the given \textit{m}th series, by following a policy $\bm{\mu}_{\bm{\theta}}$, allows to construct the cumulative objective in \eqref{seq:policy_search_objective} used for training. 
Additionally, and thanks to the closed-form policy evaluation discussed in Sec. \ref{ssec:policy_evaluation}, computing and propagating the gradient of the \textit{t}th action $\bm{a}_t$ with respect to the parameters contained in $\bm{\theta}$ is done avoiding the need of unrolling numerical optimizers \cite{monga2021algorithm} or using specific numerical tools for DCOLs such as CVXPY Layers \cite{agrawal2019differentiable}.

    % ALTERNATIVE METHODS
\section{Benchmark and baseline methods} \label{sec:alternative}
Recall from Sec. \ref{ssec:smoothing_spline_interpolation} that the batch formulation provides the optimal reconstruction with hindsight.
The batch solution can be found by solving the optimization problem \eqref{eq:smoothing_interpolation}, but only once all time-series data are available.
Thus, it cannot be used for zero-delay interpolation.
Conceptually, online methods achieve a zero-delay response at the expense of incurring higher or equal loss than the batch solution.
For this reason, the batch solution is used here as a baseline.

On the other hand, as stated in the \hyperref[sec:introduction]{Introduction} and to the best of our knowledge, there is no related work to our trainable zero-delay smoothing interpolation approach in the literature.
One could consider that the closest approach is the interpolation method known as myopic.
This is a local method in the sense that it only focuses on the last received data sample while completely ignoring the distribution of future arriving data.
For this reason, the myopic method is used here as a benchmark.
In this sense, our proposed method must outperform the myopic method to be deemed acceptable.

% MYOPIC BENCHMARK
\subsection{Myopic benchmark} \label{ssec:myopic_benchmark}
A policy that chooses the action that minimizes the current or instantaneous cost is commonly referred to as myopic.
It can be constructed as
\begin{equation} \label{eq:myopic_policy}
    \bm{\mu}(\bm{s}_t) = \text{arg}\underset{\bm{a}\in\mathcal{A}(\bm{s}_t)}{\text{min}} \left\{ \kappa(\bm{s}_t,\bm{a}) \right\} ,
\end{equation}
with cost $\kappa$ as in \eqref{eq:cost} and admissible action set as described in Sec. \ref{ssec:action_space}. 
Notice that since the myopic policy does not contain trainable parameters, it does not need to be trained.
Moreover, the myopic approach is carried out as a parameterless CFA-based policy, hence, becoming a particular case of \eqref{eq:parametric_policy}. 
For this reason, it also admits a unique and closed-form evaluation.
See Appendix \ref{apx:closed-form_policy_evaluation} for more details.
    % EXPERIMENTS
\section{Experiments} \label{sec:experiments}
%%% Figure T-V curve
\begin{figure*}
\begin{subfigure}{\textwidth}
    \centering
    \includegraphics[width=0.7\linewidth]{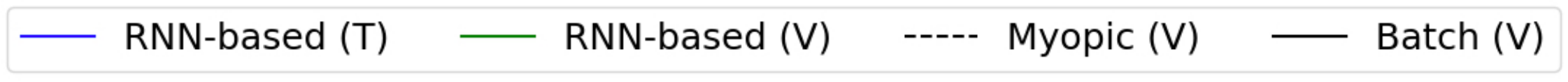}
\end{subfigure}
\hfill
\begin{subfigure}{.33\textwidth}
    \centering
    \includegraphics[width=\linewidth]{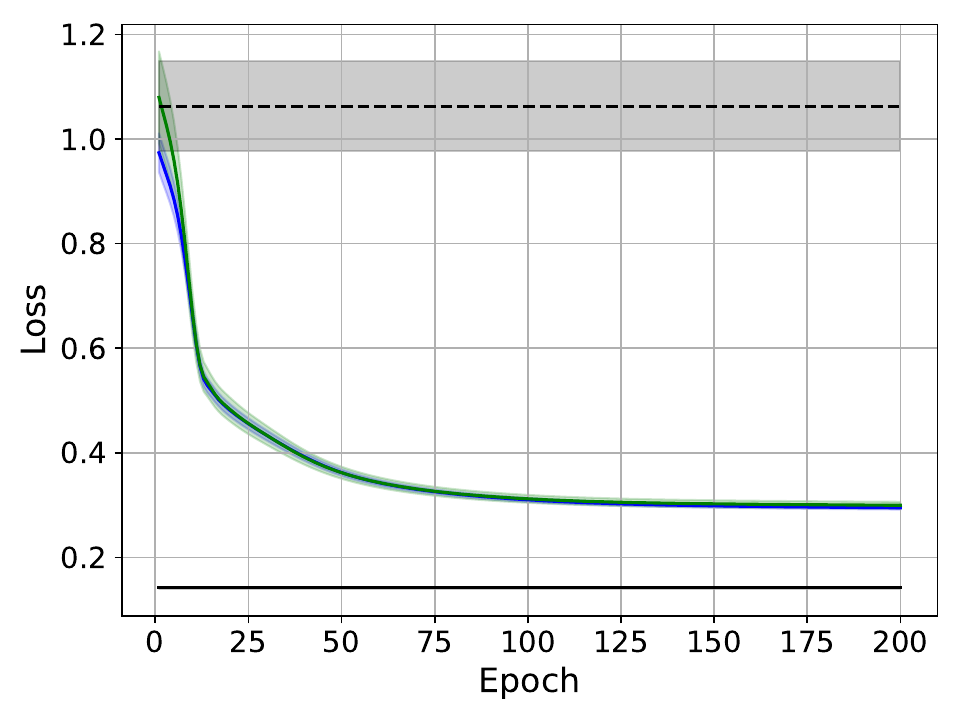}
    \caption{(4,2), $\eta=10$, and synthetic dataset.}
    \label{subfig:S_4-2}
\end{subfigure}
\begin{subfigure}{.33\textwidth}
    \centering
    \includegraphics[width=\linewidth]{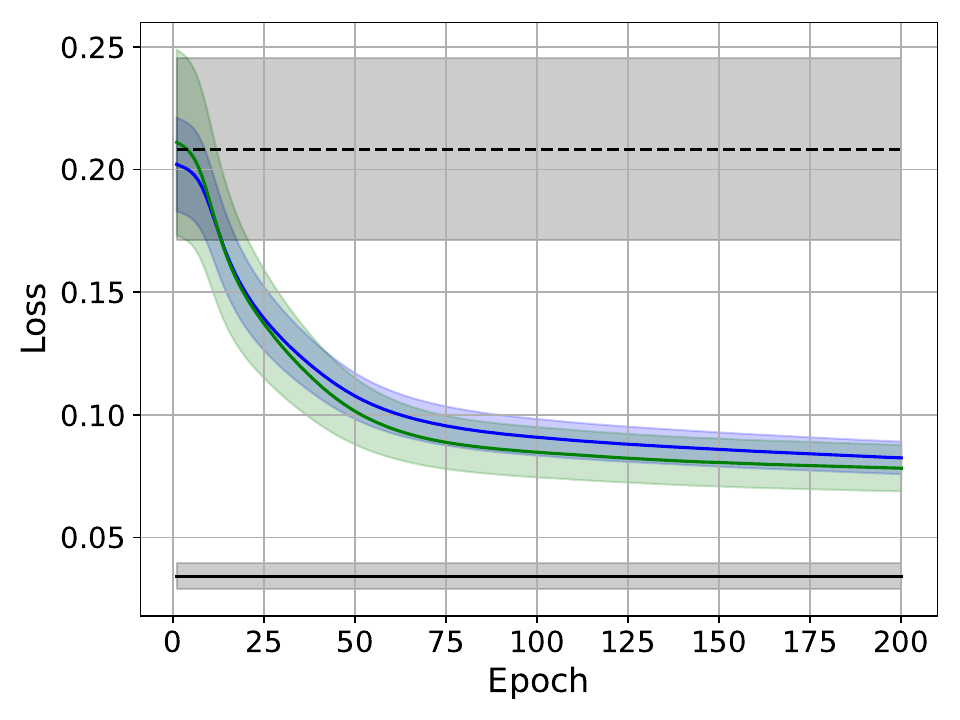}
    \caption{(3,1), $\eta=1$, and R1.}
    \label{subfig:R1_3-1}
\end{subfigure}
\begin{subfigure}{.33\textwidth}
    \centering
    \includegraphics[width=\linewidth]{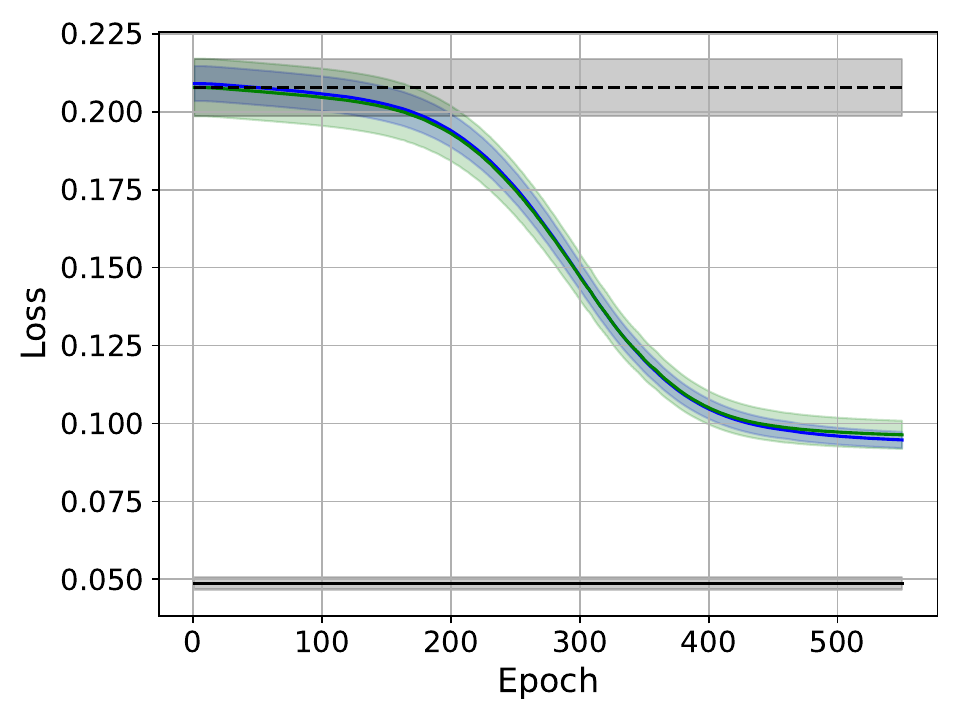}
    \caption{(4,2), $\eta=10$, and R2.}
    \label{subfig:R2_3-1}
\end{subfigure}
\hfill\par
\begin{subfigure}{.33\textwidth}
    \centering
    \includegraphics[width=\linewidth]{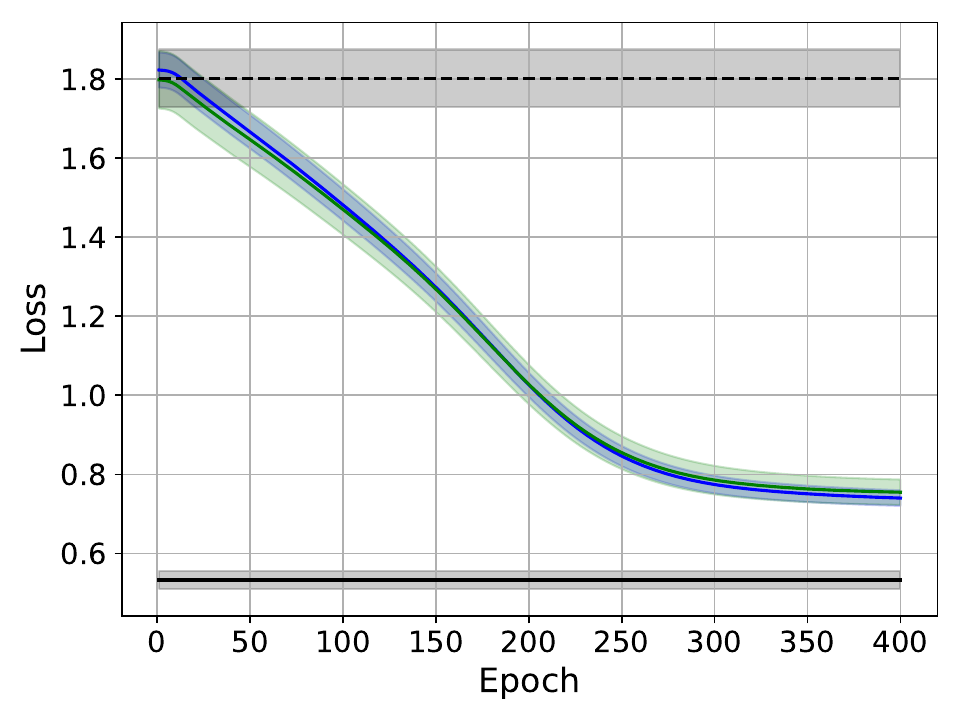}
    \caption{(4,2), $\eta=1$, and R3.}
    \label{subfig:R3_4-2}
\end{subfigure}
\begin{subfigure}{.33\textwidth}
    \centering
    \includegraphics[width=\linewidth]{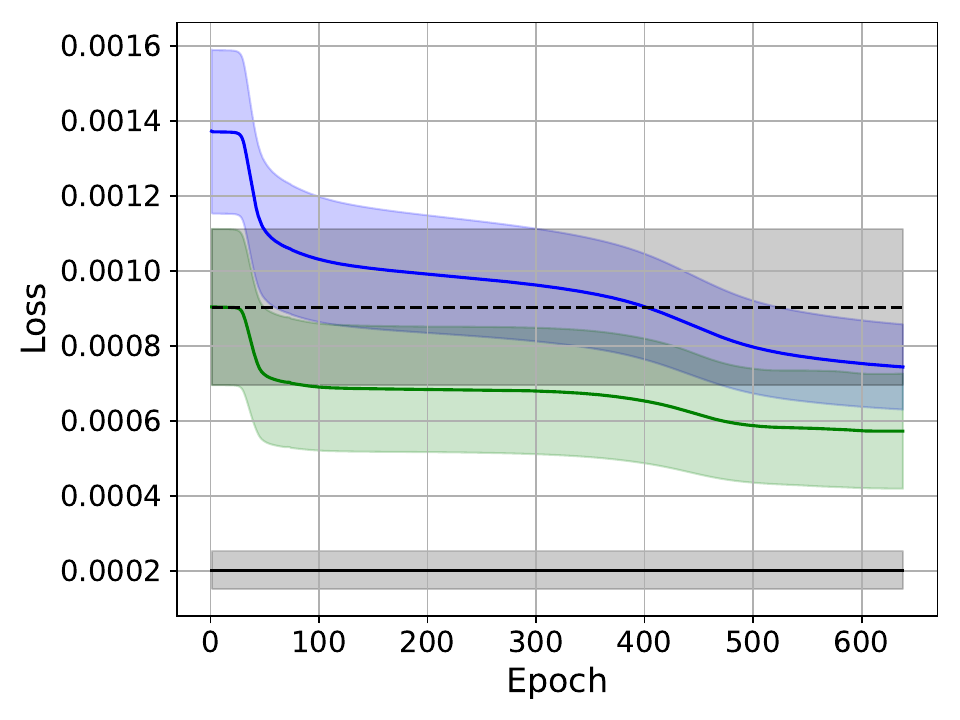}
    \caption{(4,2), $\eta=1$, and R4.}
    \label{subfig:R4_3-1}
\end{subfigure}
\begin{subfigure}{.33\textwidth}
    \centering
    \includegraphics[width=\linewidth]{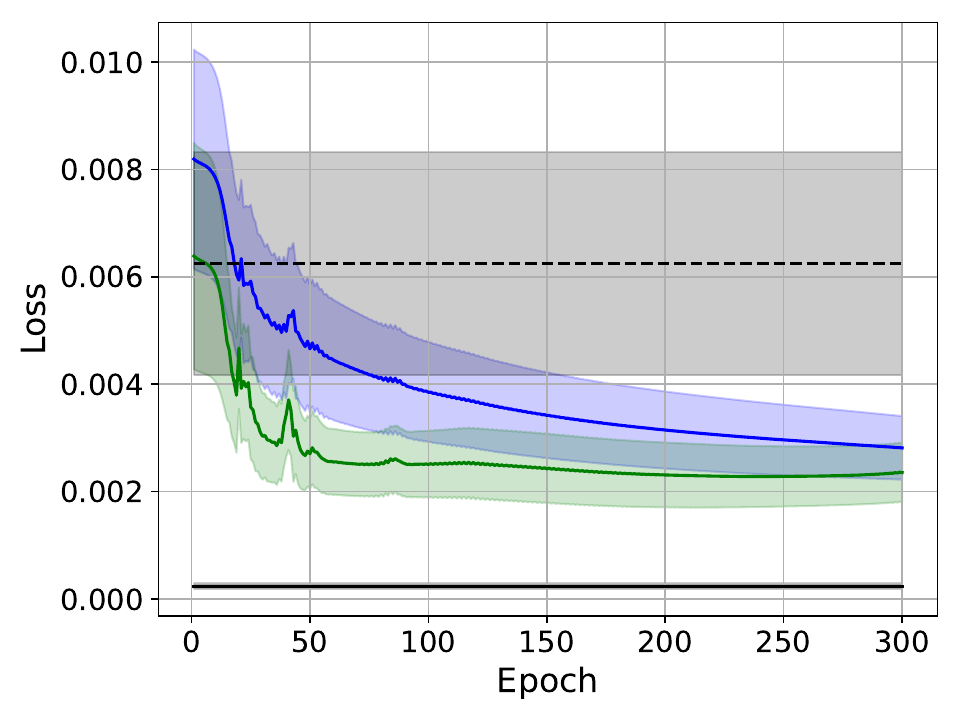}
    \caption{(3,1), $\eta=10$, and R5.}
    \label{subfig:R5_3-1}
\end{subfigure}
\caption{Some of the training-validation curves for the considered RNN-based policy configurations $(d,\varphi)$ and $\eta$ values for each dataset. The legends (T) and (V) refer to the training and validation partitions, respectively. 
The loss metric is the average total cost per function section (see Sec. \ref{ssec:total_expected_cost}). The shaded areas represent one standard deviation.}
\label{fig:T-V_curves}
\end{figure*}
%%% Table
\begin{table*}
\caption{Performance metrics averaged over the test partitions, for different $\eta$ values and policy configurations with $\rho=2$.
Recall that the improvement metric \eqref{eq:improvement} reports the gain of any RNN$(d,\varphi)$ configuration over its corresponding Myopic$(d,\varphi)$ (as a benchmark) and batch$(d,\varphi)$ (as the baseline) configurations.} 
\centering
\setlength\tabcolsep{5.2pt} % default 6pt
\renewcommand{\arraystretch}{1.25}
\begin{tabular}{|c||c||c|c|c||c|c|c||c|c|c|}
\cline{3-11}
\multicolumn{2}{c|}{} & \multicolumn{3}{c||}{$\eta=0.1$} & \multicolumn{3}{c||}{$\eta=1$} & \multicolumn{3}{c|}{$\eta=10$} \\
\cline{3-11} \noalign{\vskip\doublerulesep \vskip-\arrayrulewidth} \hline 
\textbf{Dataset} & \textbf{Configuration} & MSE & MAE & \textbf{Improvement} & MSE & MAE & \textbf{Improvement} & MSE & MAE & \textbf{Improvement} \\ 
%% S
\hline\hline
\multirow{6}{*}{Synthetic}  & Myopic$(3,1)$ & 0.49 & 0.60 & \multirow{3}{*}{$71.5\%\pm2.8\%$} & 0.31 & 0.46 & \multirow{3}{*}{$51.1\%\pm3.9\%$} & 0.64 & 0.63 & \multirow{3}{*}{$78.1\%\pm2.5\%$}   \\ 
\cline{2-4} \cline{6-7} \cline{9-10} & Batch$(3,1)$ & 0.41 & 0.56 &  & 0.26 & 0.44 &  & 0.22 & 0.39 &  \\ 
\cline{2-4} \cline{6-7} \cline{9-10} & RNN$(3,1)$ & 0.29 & 0.47 & & 0.25 & 0.42 & & 0.28 & 0.43 & \\
\cline{2-11} & Myopic$(4,2)$ & 0.50 & 0.60 & \multirow{3}{*}{$64.8\%\pm2.6\%$} & 0.38 & 0.50 & \multirow{3}{*}{$53.0\%\pm3.6\%$} & 0.83 & 0.71 & \multirow{3}{*}{$81.0\%\pm2.6\%$} \\
\cline{2-4} \cline{6-7} \cline{9-10} & Batch$(4,2)$ & 0.41 & 0.56 &  & 0.26 & 0.44 &  & 0.22 & 0.39 &  \\
\cline{2-4} \cline{6-7} \cline{9-10} & RNN$(4,2)$ & 0.28 & 0.44 &  & 0.27 & 0.43 &  & 0.30 & 0.44 &  \\
%% R1
\hline\hline
\multirow{6}{*}{R1}  & Myopic$(3,1)$ & 0.09 & 0.13 & \multirow{3}{*}{$59.4\%\pm20.3\%$} & 0.20 & 0.22 & \multirow{3}{*}{$81.3\%\pm8.2\%$} & 0.44 & 0.40 & \multirow{3}{*}{$75.5\%\pm8.2\%$}   \\
\cline{2-4} \cline{6-7} \cline{9-10} & Batch$(3,1)$ & 0.06 & 0.11 &  & 0.07 & 0.12 &  & 0.10 & 0.16 &  \\ 
\cline{2-4} \cline{6-7} \cline{9-10} & RNN$(3,1)$ & 0.07 & 0.12 &  & 0.10 & 0.15 &  & 0.19 & 0.24 &  \\
\cline{2-11} & Myopic$(4,2)$ & 0.13 & 0.16 & \multirow{3}{*}{$59.6\%\pm17.0\%$} & 0.29 & 0.27 & \multirow{3}{*}{$78.6\%\pm8.6\%$} & 0.51 & 0.43 & \multirow{3}{*}{$72.9\%\pm8.5\%$} \\
\cline{2-4} \cline{6-7} \cline{9-10} & Batch$(4,2)$ & 0.06 & 0.11 &  & 0.07 & 0.12 &  & 0.10 & 0.16 &  \\
\cline{2-4} \cline{6-7} \cline{9-10} & RNN$(4,2)$ & 0.08 & 0.13 &  & 0.12 & 0.17 &  & 0.23 & 0.27 &  \\
%% R2
\hline\hline
\multirow{6}{*}{R2}  & Myopic$(3,1)$ & 0.42 & 0.50 & \multirow{3}{*}{$67.3\%\pm6.4\%$} & 0.40 & 0.48 & \multirow{3}{*}{$52.2\%\pm7.1\%$} & 0.32 & 0.42 & \multirow{3}{*}{$68.8\%\pm5.7\%$}   \\ 
\cline{2-4} \cline{6-7} \cline{9-10} & Batch$(3,1)$ & 0.27 & 0.40 &  & 0.26 & 0.39 &  & 0.23 & 0.37 &  \\ 
\cline{2-4} \cline{6-7} \cline{9-10} & RNN$(3,1)$ & 0.23 & 0.36 &  & 0.23 & 0.37 &  & 0.20 & 0.34 &  \\
\cline{2-11} & Myopic$(4,2)$ & 0.65 & 0.62 & \multirow{3}{*}{$70.1\%\pm3.9\%$} & 0.58 & 0.59 & \multirow{3}{*}{$52.0\%\pm5.4\%$} & 0.39 & 0.47 & \multirow{3}{*}{$73.5\%\pm4.0\%$} \\
\cline{2-4} \cline{6-7} \cline{9-10} & Batch$(4,2)$ & 0.27 & 0.40 &  & 0.26 & 0.39 &  & 0.23 & 0.37 &  \\
\cline{2-4} \cline{6-7} \cline{9-10} & RNN$(4,2)$ & 0.21 & 0.35 &  & 0.22 & 0.36 &  & 0.21 & 0.35 &  \\
%% R3
\hline\hline
\multirow{6}{*}{R3}  & Myopic$(3,1)$ & 4.60 & 1.80 & \multirow{3}{*}{$76.3\%\pm4.4\%$} & 2.90 & 1.40 & \multirow{3}{*}{$81.3\%\pm5.3\%$} & 1.80 & 1.09 & \multirow{3}{*}{$65.6\%\pm11.6\%$}   \\ 
\cline{2-4} \cline{6-7} \cline{9-10} & Batch$(3,1)$ & 3.10 & 1.40 &  & 2.35 & 1.30 &  & 1.66 & 1.06 &  \\
\cline{2-4} \cline{6-7} \cline{9-10} & RNN$(3,1)$ & 2.56 & 1.30 &  & 1.56 & 1.02 &  & 1.40 & 0.96 &  \\
\cline{2-11} & Myopic$(4,2)$ & 6.30 & 2.10 & \multirow{3}{*}{$69.7\%\pm4.4\%$} & 3.02 & 1.43 & \multirow{3}{*}{$88.9\%\pm3.9\%$} & 1.89 & 1.10 & \multirow{3}{*}{$80.8\%\pm9.0\%$} \\
\cline{2-4} \cline{6-7} \cline{9-10} & Batch$(4,2)$ & 3.10 & 1.40 &  & 2.36 & 1.26 &  & 1.66 & 1.06 &  \\
\cline{2-4} \cline{6-7} \cline{9-10} & RNN$(4,2)$ & 1.80 & 1.00 &  & 1.43 & 0.95 &  & 1.52 & 1.00 &  \\
%% R4
\hline\hline
\multirow{6}{*}{R4} & Myopic$(3,1)$ & 4.5e-3 & 2.8e-2 & \multirow{3}{*}{$13.7\%\pm57.3\%$} & 4.1e-3 & 2.7e-2 & \multirow{3}{*}{$-61.8\%\pm94.4\%$} & 3.6e-3 & 2.5e-2 & \multirow{3}{*}{$16.5\%\pm54.5\%$}   \\ 
\cline{2-4} \cline{6-7} \cline{9-10} & Batch$(3,1)$ & 2.7e-3 & 2.3e-2 &  & 2.6e-3 & 2.2e-2 &  & 2.4e-3 & 2.0e-2 &  \\ 
\cline{2-4} \cline{6-7} \cline{9-10} & RNN$(3,1)$ & 7.1e-3 & 6.8e-2 &  & 6.0e-3 & 5.6e-2 &  & 3.2e-3 & 3.1e-2 &  \\
\cline{2-11} & Myopic$(4,2)$ & 6.9e-3 & 3.6e-2 & \multirow{3}{*}{$-35.5\%\pm78.5\%$} & 5.7e-3 & 3.2e-2 & \multirow{3}{*}{$65.7\%\pm25.5\%$} & 4.9e-3 & 3.0e-2 & \multirow{3}{*}{$19.5\%\pm51.5\%$} \\
\cline{2-4} \cline{6-7} \cline{9-10} & Batch$(4,2)$ & 2.7e-3 & 2.3e-2 &  & 2.6e-3 & 2.2e-2 &  & 2.4e-2 & 2.0e-2 &  \\
\cline{2-4} \cline{6-7} \cline{9-10} & RNN$(4,2)$ & 3.7e-3 & 3.3e-2 &  & 4.1e-3 & 3.1e-2 &  & 3.5e-3 & 2.7e-2 &  \\
%% R5
\hline\hline
\multirow{6}{*}{R5} & Myopic$(3,1)$ & 2.7e-4 & 9.3e-3 & \multirow{3}{*}{$-55.4\%\pm101.6\%$} & 6.0e-4 & 1.3e-2 & \multirow{3}{*}{$26.3\%\pm59.8\%$} & 7.1e-3 & 3.8e-2 & \multirow{3}{*}{$88.2\%\pm7.1\%$}   \\ 
\cline{2-4} \cline{6-7} \cline{9-10} & Batch$(3,1)$ & 1.8e-4 & 7.8e-3 &  & 1.7e-4 & 7.7e-3 &  & 2.6e-4 & 9.1e-3 &  \\ 
\cline{2-4} \cline{6-7} \cline{9-10} & RNN$(3,1)$ & 2.3e-4 & 8.7e-3 &  & 3.1e-4 & 1.0e-2 &  & 3.2e-3 & 3.0e-2 &  \\
\cline{2-11} & Myopic$(4,2)$ & 4.7e-4 & 1.1e-2 & \multirow{3}{*}{$-13.3\%\pm68.2\%$} & 9.7e-4 & 1.7e-2 & \multirow{3}{*}{$39.8\%\pm41.9\%$} & 1.1e-2 & 4.8e-2 & \multirow{3}{*}{$85.1\%\pm7.5\%$} \\
\cline{2-4} \cline{6-7} \cline{9-10} & Batch$(4,2)$ & 1.8e-4 & 7.8e-3 &  & 1.7e-4 & 7.7e-3 &  & 2.6e-4 & 9.1e-3 &  \\
\cline{2-4} \cline{6-7} \cline{9-10} & RNN$(4,2)$ & 2.9e-4 & 9.7e-3 &  & 6.1e-4 & 1.4e-2 &  & 5.4e-3 & 3.6e-2 &  \\
\hline
\end{tabular}
\renewcommand{\arraystretch}{1}
\label{table:improvement}
\end{table*}

In this section, we experimentally validate the effectiveness of the proposed RNN-based policy, introduced in Sec. \ref{sec:solution}.
To this end, we first describe the time-series datasets used.
Then, we outline the possible policy configurations, i.e., the possible types of splines as well as the RNN architecture.
Afterward, we report how the experiments have been carried out.
Finally, we present and comment on the experimental results.

%% DESCRIPTION OF PROBLEM DATA
\subsection{Problem data description} \label{ssec:problem_data_description}
For these experiments, we use a synthetic dataset and five real datasets. 
Each dataset consists of a time series of 28800 signal samples which has been split into 288 series of 100 samples each, except for the first real dataset which contains 57600 samples split into 576 series.

The synthetic dataset is first generated as a uniformly arranged realization of a given autoregressive process AR(2) with white Gaussian noise $\mathcal{N}(0,0.1)$.
Then, the resulting series is compressed via PI \cite{softpi} with $\text{CompDev}=0.1$, $\text{CompMax}=\infty$ and $\text{CompMin}=0$.
As a result, the series time stamps are not uniformly distributed anymore.

The first real dataset (R1) consists of a series of household minute-averaged active power consumption (in Kilowatts) \cite{misc_individual_household_electric_power_consumption_235}. 
The second real dataset (R2) is a quantized and PI-compressed (and hence not uniformly sampled) time series measuring an oil separation deposit pressure\footnote{Data collected from Lundin's offshore oil and gas platform Edvard-Grieg.} (in Bar).
For the third real dataset (R3) \cite{ScalabriniSampaio2019}, a cooling fan with weights on its blades is used to generate vibrations which are recorded by an attached accelerometer.
The vibration samples are recorded every $20$ milliseconds.
We use the accelerometer recorded $x$-values (which are standardized) for the rotation speeds ranging from 5 to 40 rpm.
The fourth real dataset (R4) \cite{skintemp} monitors the skin temperature (in Celsius degrees) of a volunteer subject through a wearable device every 4 minutes.
The fifth and last real dataset (R5) \cite{bodik2004intel} consists of a sensor within a sensor network deployed in a lab, collecting the temperature-corrected relative humidity in percentage. 
The sampling rate is non-uniform and ranges from deciseconds to tens of seconds. 
Finally, it is worth mentioning that the datasets R4 and R5 contain gaps (several orders of magnitude wider than the average sampling period) of missing data that we have shortened to avoid instability in the reconstruction.
In similar cases where the available raw data is of low quality, thorough and task-specific data preprocessing techniques are assumed.
This can improve the performance results as described in the ensuing Sec. \ref{ssec:results_discussion}.

%% POLICY CONFIGURATION
\subsection{Policy configuration} \label{ssec:policy_configuration}
We experimentally observe that the myopic policy described in Sec. \ref{ssec:myopic_benchmark} is not stable for values of $\rho>2$.
Recall that the value of $\rho$ affects the policy cost, set as in \eqref{eq:cost}, and delimits the order and degree of smoothness of the spline signal estimate, as explained in Sec. \ref{ssec:splines}. 
We also observe instability under the myopic policy for $\rho=2$ with a spline signal estimate of order $d=3$ and degree of smoothness $\varphi=2$.
Consequently, our proposed RNN-based policy is unstable for the same $\rho$ values and spline configurations since it implicitly uses the myopic policy as a guided starting point. 
This can be seen by comparing \eqref{eq:myopic_policy} and \eqref{eq:parametric_policy} with a near-zero initial value of $\lambda$.
Although further theoretical instability studies, alternative policy architectures, or low-delay approaches can contribute to solving the instability issue, they lie outside of the scope of this paper. 
Nonetheless, we have maintained the general problem formulation as a starting point for future works to take over.
On the other hand, the interpolation problem with $\rho=1$ is not interesting since it leads to linear interpolation.
Therefore, in the present work, we focus on the smoothing interpolation problem with $\rho = 2$ and with the remaining stable spline configurations, within the search function space described in Sec. \ref{ssec:smoothing_spline_interpolation}, which can lead to optimal reconstructions. 
Those spline configurations, hereinafter specified by the shorthand notation $(d,\varphi)$ of the order and degree of smoothness of the spline, correspond to (3,1) and (4,2).
Accordingly, the notation Myopic$(d,\varphi)$ or RNN$(d,\varphi)$ refers to the type of policy besides the spline configuration.

Regarding the RNN architecture shaping the approximated cost-to-go within the RNN-based policy, introduced in Sec. \ref{ssec:policy_form} and illustrated in Fig. \ref{fig:rnn_architecture}, we set a preprocessing step that forwards the time length, i.e. $u_t=x_t - x_{t-1}$, of the \textit{t}th time section $\mathcal{T}_t$, instead of directly using the time stamps.
This preprocessing step makes the architecture invariant to time shifts in the set of time stamps.
In our experiments, the recurrent unit consists of two stacked gated recurrent unit (GRU) layers \cite{cho2014properties}, with a latent state (hidden state) of size 16 and an input of size 16. 
The input and output layers are set as linear layers to match the required dimensionality, i.e., to match the input size after the preprocessing step and to match the order of the spline minus the number of constrained coefficients as output size.

%% EXPERIMENTAL SETUP
\subsection{Experimental setup}
The datasets are randomly divided into 192 series for training, 64 for validation, and 32 for testing. 
Except for the R1 dataset, which has been divided in the same proportion but in relation to its data size.
The benefits of this train-validation-test partition are two-fold: i) the policy becomes more robust against unknown initial conditions, and ii) we can validate the reconstruction against an optimal batch solution (shorter sequences are computationally tractable using batch optimization).
All series within a dataset are standardized for implementation convenience.
To avoid data leaking, the mean and standard deviation of their respective training partition are used for the standardization.
In other words, we compute the mean and variance of the training partition and assume them to be the moments of the true data distribution.
The standardization of series is useful to enforce the RNN unit to focus on the fluctuations of the signal values rather than on their magnitude.
Finally, the RNN-based policy has been trained using the adaptive moments (Adam) optimizer~\cite{kingma2014adam}, with $\beta_1 = 0.9$, $\beta_2 = 0.999$, without weight decay, and a learning rate of $0.001$ over mini-batches of 32 time series each (double mini-batch size in the case of R1).

%% RESULTS AND DISCUSSION
\subsection{Results and discussion} \label{ssec:results_discussion}
%%% Figure lambda training curves
\begin{figure}
    \centering
    \includegraphics[width=0.98\columnwidth]{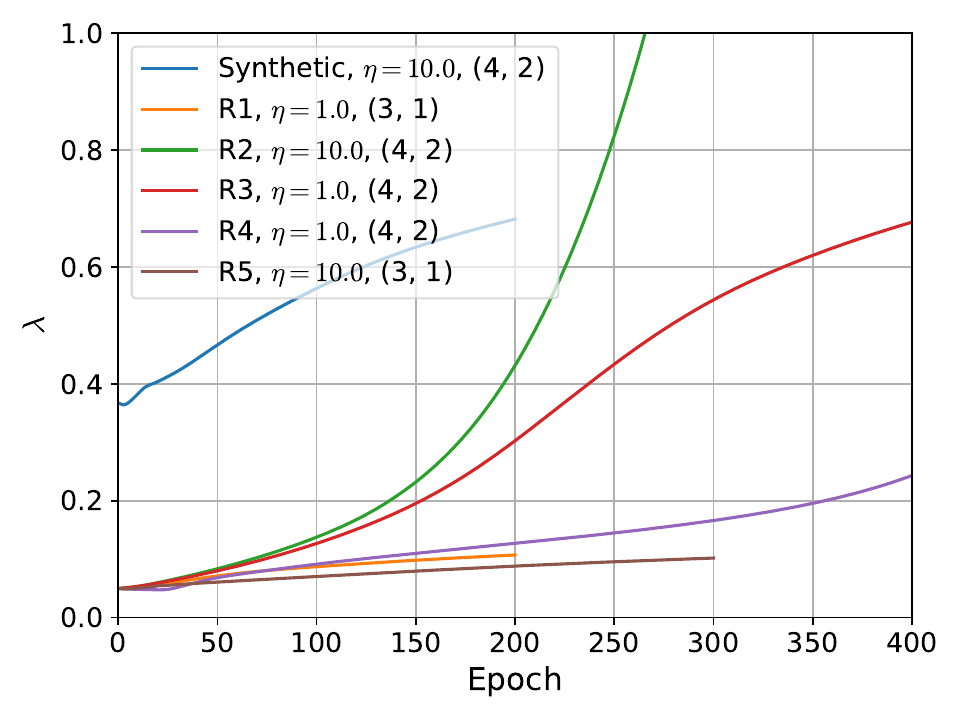}
    \caption{Training curves of the parameter $\lambda\in\mathbb{R}_+$ introduced in \eqref{eq:cost-to-go_approximation}. The elements displayed coincide with those shown in Fig. \ref{fig:T-V_curves}.}
    \label{fig:lambda_weights}
\end{figure}
%%% Figure FPtime
\begin{figure}
    \centering
    \includegraphics[width=0.975\columnwidth]{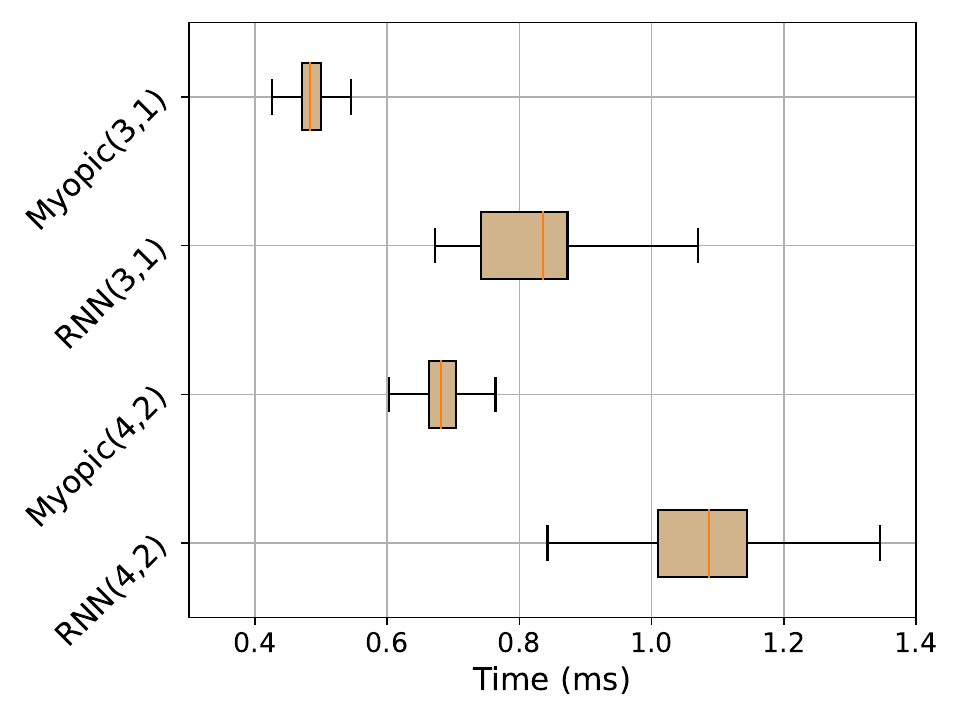}
    \caption{Box plot of the policy execution time per interpolation step over the test partitions and $\eta$ values $\{0.1,1,10\}$. 
    It illustrates (excluding outliers) the minimum, first quartile, median, third quartile, and maximum.}
    \label{fig:fp_boxplot}
\end{figure}

Some of the training-validation curves are presented in Fig. \ref{fig:T-V_curves}.
As expected, we observe that randomly initialized RNN-based policies (except for the parameter $\lambda$, which controls the length of the initial performance gap, as discussed in Sec. \ref{ssec:policy_configuration}, and is manually initialized) only outperform the myopic policy after training.
We also observe wider (in relative terms) standard deviations in those datasets with more abrupt changes, either from the nature of the data, as in R1, or due to missing data and posterior preprocessing, as in the case of R4 and R5.
This phenomenon appears also to be caused by highly non-uniform sampling rates, as in R5.
But in this case, the width seems to decrease as the policy yields more accurate estimates.
This implies that the RNN-based policy is able to learn how to adapt under non-uniform sampling rates properly.

Once the RNN-based policy has been trained, we measure its performance with respect to the myopic policy (as the benchmark) and the batch reconstruction (as the baseline) through an \emph{improvement} metric defined as
\begin{equation} \label{eq:improvement}
    I = \frac{\ell_M - \ell_R}{\ell_M - \ell_B} ,
\end{equation}
where $\ell_M$, $\ell_R$, $\ell_B$ denote the loss metric displayed in Fig. \ref{fig:T-V_curves} but over the test partition for the myopic, the RNN-based policies and the batch solution, respectively.
Acceptable performances yield improvement values in $(0,1]$, being $I=1$ the best possible value, whereas nonpositive improvement values indicate a deficient performance.
The standard deviation of the improvement metric is then estimated through error propagation, i.e.,
\begin{equation}
    \sigma_I = \sqrt{ \left(\frac{\partial I}{\partial \ell_M}\right)^2\sigma^2_M + \left(\frac{\partial I}{\partial \ell_R}\right)^2\sigma^2_R + \left(\frac{\partial I}{\partial \ell_B}\right)^2\sigma^2_B } ,
\end{equation}
with $\sigma_M$, $\sigma_R$, and $\sigma_B$ denoting the standard deviation of the respective loss metrics over the test partition.
The improvement results are summarized in Table \ref{table:improvement}.
From Fig. \ref{fig:T-V_curves} and Table \ref{table:improvement}, it can be observed that the policy configurations with the highest improvement scores over each of the considered dataset test partitions are in agreement with their corresponding validation curves.
Table \ref{table:improvement} also shows standard performance descriptors such as the mean squared error (MSE) and mean absolute error (MAE). 
See Appendix \ref{apx:bootstrap} for their computation. 
Note that for most of the experiments that we have carried out, the RNN-based policy outperforms, in terms of the MSE and MAE metrics, the myopic policy while it falls behind the batch policy.
This observation experimentally justifies the smoothness assumption in our formulation.

Regarding the parameter $\lambda\in\mathbb{R}_+$ introduced in \eqref{eq:cost-to-go_approximation}, it can be understood as the confidence of the RNN-based policy in its ability to foresee incoming data samples.
In this way, it also quantifies the importance of the RNN architecture (detailed in Fig. \ref{fig:rnn_architecture}) in the reconstruction task.
As an illustration, Fig. \ref{fig:lambda_weights} shows the training curves corresponding to the parameter $\lambda$ for the policy configurations presented in Fig. \ref{fig:T-V_curves}.

On the other hand, we observe a competitive performance in terms of the execution time of the RNN-based policy evaluation (forward pass) as compared to their myopic counterpart.
Our evaluation time results are summarized in Fig. \ref{fig:fp_boxplot}, where the policies are implemented in Python 3.8.8. and the experiment is done in a 2018 laptop with a 2.7 GHz Quad-Core Intel Core i7 processor and 16 GB 2133 MHz LPDDR3 memory.
Regarding memory complexity, the myopic policy is parameterless (see Sec. \ref{ssec:myopic_benchmark}), and our configuration of the RNN-based policy (see Sec. \ref{ssec:policy_configuration}) contains approximately 3400 trainable parameters, which is arguably a reduced model size for most tasks.

\begin{figure}
    \centering
    \includegraphics[width=\columnwidth]{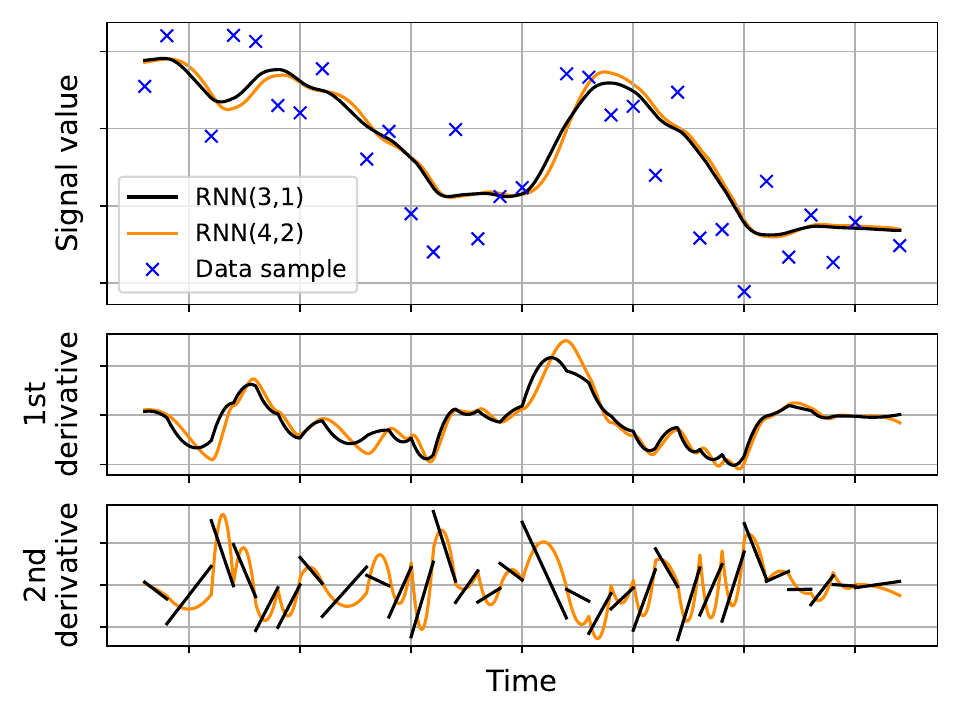}
    \caption{Snapshot of a reconstructed time series from the test partition of the synthetic dataset. 
    Both of the presented policies have been trained over the training partition before reconstruction.}
    \label{fig:example}
\end{figure}

Finally, and for the sake of completeness, Fig. \ref{fig:example} shows a snapshot of a zero-delay smooth signal reconstruction alongside its two first derivatives using our proposed method.

    % CONCLUSION
\section{Conclusion} \label{sec:conclusion}
In this paper, we propose a method for zero-delay smoothing spline interpolation.
Our method relies on a parametric policy, named the RNN-based policy, specifically engineered for the zero-delay interpolation task.
As new data samples arrive, this policy yields piecewise polynomial functions used for smooth signal reconstruction. 
Our experiments show that the RNN-based policy can learn the dynamics of the target signal and efficiently incorporate them (in terms of improved accuracy and reduced response time) into the reconstruction task.

This work can be seen as a proof of concept with several immediate follow-ups.
The flexibility in our policy design allows extending this work to multivariate time series with a moderate increase in complexity.
It is also possible to generalize the problem data, e.g., quantization intervals instead of data points, as well as to accommodate additional constraints as long as the convexity of the policy evaluation problem is preserved.
Lastly, we notice that our work provides the foundation and can be tailored effectively for reconstructing non-stationary signals by borrowing reinforcement learning techniques.

    % APPENDIX
\appendix 

%% proof proposition continuity equality constraints
\subsection{Proof of \textbf{Proposition \ref{prp:continuity}}} \label{apx:continuity}
Recall from Sec. \ref{ssec:splines} that every spline $f_{T}$, as in \eqref{eq:spline}, is composed of $T$ function sections and $T-1$ contact points.
We say that two consecutive function sections have a contact of order $\varphi$ if they have $\varphi$ equal derivatives at the contact point.
Then, guaranteeing a degree of smoothness $\varphi$ for a given spline $f_{T}$ is equivalent to ensuring that all its contact points are at least of order $\varphi$ since every \textit{t}th function section $g_t$, as in \eqref{eq:function_section}, is already smooth over the interior of its domain $\mathcal{T}_t$.
In practice, this can be ensured by imposing the following equality constraints
\begin{equation} \label{eq:continuity_constraints}
\underset{x \to x_{t-1}^-}{\text{lim }} D^k_x \, g_{t-1}(x) = \underset{x \to x_{t-1}^+}{\text{lim }} D^k_x \, g_t(x),
\end{equation}
for every $k\in \mathbb{N}^{[0,\varphi]}$ and $t\in\mathbb{N}^{[2,T]}$.
From here, notice that the $k$th derivative of every \textit{t}th function section $g_t$ can be computed as
\begin{equation} \label{eq:derivative_function_section}
    D^k_x \, g_t(x) = {\bm{a}_t}^\top \left[ D^k_x\left[\bm{p}_t(x)\right]_1, \dots, D^k_x\left[\bm{p}_t(x)\right]_{d+1} \right]^\top \, .
\end{equation}
Also, notice from the definition in \eqref{eq:basis_vector} that the \textit{i}th component of the \textit{t}th basis vector function $\bm{p}_t$ equals
\begin{equation} 
    \left[\bm{p}_t(x)\right]_i = (x - x_{t-1})^{i-1},
\end{equation}
for all $i\in[1,d+1]$.
From this point, the \textit{k}th derivative of each \textit{i}th component of the basis vector function $\bm{p}_t$ can be straightforwardly computed as
\begin{equation} \label{eq:derivative_basis_vector_element}
    D^k_x \left[\bm{p}_t(x) \right]_i = (x-x_{t-1})^{i-1-k} \prod^k_{j=1} (i-j) \, .
\end{equation}

Now observe that
\begin{equation} \label{eq:lim_derivative_basis_vector_element}
    \underset{x \to x_{t-1}^+}{\text{lim }} D^k_x \left[\bm{p}_t(x) \right]_i = 
    \begin{cases}
        k! & \text{if } i = k+1, \\
        0 & \text{otherwise} .
    \end{cases}
\end{equation}
Therefore, from the relations in \eqref{eq:derivative_function_section} and \eqref{eq:lim_derivative_basis_vector_element}, the right hand term in \eqref{eq:continuity_constraints} can be equivalently computed as
\begin{subequations}  \label{eq:derivation_eqc_rhs}
    \begin{align}
        \underset{x\to x_{t-1}^+}{\text{lim }} D^k_x \, g_t(x) &= \sum^{d+1}_{i=1} \left[ \bm{a}_t \right]_i \underset{x\to x_{t-1}^+}{\text{lim }} D^k_x \left[\bm{p}_t(x) \right]_i  \\
        &= k! \, \left[ \bm{a}_t \right]_{k+1} .
    \end{align}
\end{subequations}

Separately, we can define a vector $\bm{e}_t \in \mathbb{R}^{\varphi + 1}$ whose components are constructed as
\begin{subequations} \label{eq:derivation_eqc_lhs}
    \begin{align}
        \left[ \bm{e}_t \right]_{k+1} &\triangleq \frac{1}{k!} \underset{x\to x_t^-}{\text{lim }} D^k_x \, g_t(x) \\
        &= \frac{1}{k!} \sum^{d+1}_{i=1} \left[ \bm{a}_t \right]_i u_t^{i-1-k}\,\prod^k_{j=1}(i-j), \label{seq:derivation_eqc}
    \end{align}
\end{subequations}
for every $k\in\mathbb{N}^{[0,\varphi]}$ and $t\in\mathbb{N}^{[1,T]}$ with $u_t \triangleq x_t - x_{t-1}$, and where the step (\ref{seq:derivation_eqc}) uses the relations described in \eqref{eq:derivative_function_section} and \eqref{eq:derivative_basis_vector_element}.
On the other hand, $\bm{e}_0$ encodes the initial boundary conditions of the reconstruction and can be set by the user in advance or calculated. 
Finally, by dividing both sides of the equality constraint in \eqref{eq:continuity_constraints} by $k!$, using the relations derived in \eqref{eq:derivation_eqc_rhs} and \eqref{eq:derivation_eqc_lhs}, and appropriately renaming the indices we obtain the \textbf{Proposition \ref{prp:continuity}}.

%% proof proposition cost additive form
\subsection{Proof of \textbf{Proposition \ref{prp:additivity}}} \label{apx:additivity}
Recall from Sec. \ref{ssec:smoothing_spline_interpolation} that the solution to the optimization problem stated in \eqref{eq:smoothing_interpolation} is a spline function in $\mathcal{W}_\rho$.
This fact allows us to reduce the search function space without loss of optimality.
In fact, we can incorporate the spline form of the solution into the objective functional as far as we ensure the required minimum degree of smoothness of the solution, for example, via \eqref{eq:continuity_constraints}.
From here, we can equivalently compute the regularization term in the objective in \eqref{eq:smoothing_interpolation} (second term) as
\begin{equation} \label{eq:roughness_relation}
    \int_{\bigcup_{t=1}^{T} \mathcal{T}_t} \left( D^\rho_x \, f_{T}(x) \right)^2 dx = \sum^{T}_{t=1} \int_{\mathcal{T}_t} \left( D^\rho_x \, g_t(x) \right)^2 dx \, .
\end{equation}

Separately, and making use of the definition of function section in \eqref{eq:function_section}, we obtain the following relation
\begin{subequations} \label{eq:roughness_function_section}
    \begin{align}
        \int_{\mathcal{T}_t} \left( D^\rho_x \, g_t(x) \right)^2 dx &= \int_{\mathcal{T}_t} \left( D^\rho_x \, \bm{a}^\top_t \bm{p}_t(x) \right)^2 dx \\
        &= \bm{a}^\top_t \bm{M}_t \bm{a}_t , 
    \end{align}
\end{subequations}
with
\begin{equation} \label{eq:basic_definition_M}
    [\bm{M}_t]_{i,j} = \int_{\mathcal{T}_t} D^\rho_x\left[\bm{p}_t(x)\right]_i \, D^\rho_x\left[\bm{p}_t(x)\right]_j \, dx \, .
\end{equation}
From the relation in \eqref{eq:derivative_basis_vector_element}, it is clear that the first $\rho$ rows and columns of the matrix defined in \eqref{eq:basic_definition_M} are zero valued.
Then, we can compute the rest of the elements in the matrix $\bm{M}_t\in\bm{S}^{d+1}_+$ as follows
\begin{subequations} \label{eq:derivation_M}
\begin{align}
        [\bm{M}_t]_{i,j} &= \prod^\rho_{k=1}(i-k)(j-k) \int^{x_t}_{x_{t-1}} (x - x_{t-1})^{i+j-2(\rho+1)} \, dx \\
        &= \frac{(x_t - x_{t-1})^{i+j-2\rho-1}}{i+j-2\rho-1} \prod^\rho_{k=1}(i-k)(j-k) \, .
\end{align}
\end{subequations}

On the other hand, the sum of squared residuals in the objective in \eqref{eq:smoothing_interpolation} (first term) can be equivalently computed as 
\begin{equation} \label{eq:squared_residuals_relation}
    \sum^{T}_{t=1} \left( f_{T}(x_t) - y_t \right)^2 = \sum^{T}_{t=1} \left( g_t(x_t) - y_t \right)^2 ,
\end{equation}
from the definition of spline, see relation \eqref{eq:spline}.

Summing up, the result stated in \textbf{Proposition \ref{prp:additivity}} can be reached starting from the objective in \eqref{eq:smoothing_interpolation} then following the relations in \eqref{eq:roughness_relation}, \eqref{eq:roughness_function_section} alongside \eqref{eq:derivation_M} and \eqref{eq:squared_residuals_relation}.

%% closed-form policy evaluation
\subsection{Closed-form policy evaluation} \label{apx:closed-form_policy_evaluation}

\begin{table}
\caption{Terms in \eqref{eq:policy_quadratic}. The notation is shared with the rest of the paper with the incorporation of $\bm{P}_t \triangleq \bm{p}_t(x_t)\bm{p}_t(x_t)^\top$ and $\bm{v}_t \triangleq [\bm{0}^\top_{\varphi+1},{\bm{r}_t}^\top]^\top$.}
\centering
\renewcommand{\arraystretch}{1.5}
\begin{tabular}{|c|c|c|}
\cline{2-3}
\multicolumn{1}{l|}{} & $\bm{A}_t$   & $\bm{b}_t$   \\  \cline{2-3} \noalign{\vskip\doublerulesep \vskip-\arrayrulewidth} \hline
\textbf{Myopic} & $\bm{P}_t + \eta \bm{M}_t$ & $-2y_t\bm{p}_t(x_t)$ \\ \hline 
\textbf{RNN} & $\bm{P}_t + \eta \bm{M}_t + \lambda \bm{I}_{d+1}$ & $-2(y_t\bm{p}_t(x_t) + \lambda\bm{v}_t)$ \\ \hline
\end{tabular}
\renewcommand{\arraystretch}{1}
\label{table:policy_quadratic}
\end{table}

Notice that both the proposed policy in \eqref{eq:parametric_policy} and the myopic policy in \eqref{eq:myopic_policy} can be equivalently evaluated by solving the following quadratic convex problem
\begin{equation} \label{eq:policy_quadratic}
    \bm{\mu}(\bm{s}_t) = \text{arg} \underset{\bm{a}\in\mathcal{A}(\bm{s}_t)}{\text{min}} \left\{  \bm{a}^\top \bm{A}_t \bm{a} + \bm{b}^\top_t\bm{a} \right\} ,
\end{equation}
where the terms $\bm{A}_t \in \bm{S}_+^{d+1}$ and $\bm{b}_t \in \mathbb{R}^{d+1}$ take different values for the different policy variations as described in the Table \ref{table:policy_quadratic}.
The form in \eqref{eq:policy_quadratic} is displayed as an intermediate step for the sake of clarity, and the dependencies with example time series (indexed by $m$) and the policy parameters (contained in $\bm{\theta}$) have been omitted for the sake of notation.
Then, we relocate the equality constraints (presented in \eqref{eq:cc_vf} and satisfied by the actions in the admissible set $\mathcal{A}(\bm{s}_t)$) in the objective of \eqref{eq:policy_quadratic}, by restating
\begin{equation}
    \bm{a} = \begin{bmatrix} \bm{e}_{t-1} \\ \bm{0}_{d-\varphi} \end{bmatrix} + \begin{bmatrix} \bm{0}_{\varphi+1} \\ \bm{\alpha} \end{bmatrix} ,
\end{equation}
or equivalently, by setting $\bm{a} = \bm{B}_1\bm{e}_{t-1} + \bm{B}_2\bm{\alpha}$ where the components of $\bm{e}_{t-1}$ are described in \eqref{eq:cc_element}, the matrices $\bm{B}_1 \triangleq [\bm{I}_{\varphi + 1}, \bm{0}_{(\varphi + 1)\times(d-\varphi)}]^\top \in \mathbb{R}^{(d+1)\times(\varphi+1)}$ and $\bm{B}_2 \triangleq [\bm{0}_{(d-\varphi)\times(\varphi+1)},\bm{I}_{d-\varphi}]^\top \in \mathbb{R}^{(d+1)\times(d-\varphi)}$ are defined for the sake of notation, and where $\bm{\alpha}\in\mathbb{R}^{d-\varphi}$.
After some algebraic steps, both policies can be equivalently evaluated as
\begin{equation}
    \bm{\mu}(\bm{s}_t) = 
    \begin{bmatrix}
    \bm{e}_{t-1} \\
    \bm{\alpha}_t
    \end{bmatrix} ,
\end{equation}
where the vector $\bm{\alpha}_t \in \mathbb{R}^{d-\varphi}$ is obtained from
\begin{subequations}
\begin{align}
    \bm{\alpha}_t = \text{arg} \underset{\bm{a}\in\mathbb{R}^{d-\varphi}}{\text{min}} \{ & \bm{\alpha}^\top\bm{B}^\top_2\bm{A}_t\bm{B}_2\bm{\alpha}  \\
    &  + 2\left( \bm{e}^\top_{t-1}\bm{B}^\top_1\bm{A}_t\bm{B}_2 + \bm{b}^\top_t\bm{B}_2\right)\bm{\alpha} \} ,
\end{align}
\end{subequations}
with closed-form solution given by
\begin{equation} \label{eq:alpha}
    \bm{\alpha}_t = - \left( \bm{B}_2^\top \bm{A}_t \bm{B}_2 \right)^{-1} \left( \bm{B}_2\bm{A}_t\bm{B}_1\bm{e}_{t-1} + \frac{1}{2}\bm{B}_2^\top\bm{b}_t \right) ,
\end{equation}
being $\bm{B}_2^\top \bm{A}_t \bm{B}_2 \in \bm{S}^{d-\varphi}_{++}$.

%% Bootstrapping
\subsection{Boostrap method for estimating the MSE and MAE} \label{apx:bootstrap}
\begin{figure}
    \centering
    \includegraphics[width=0.875\columnwidth]{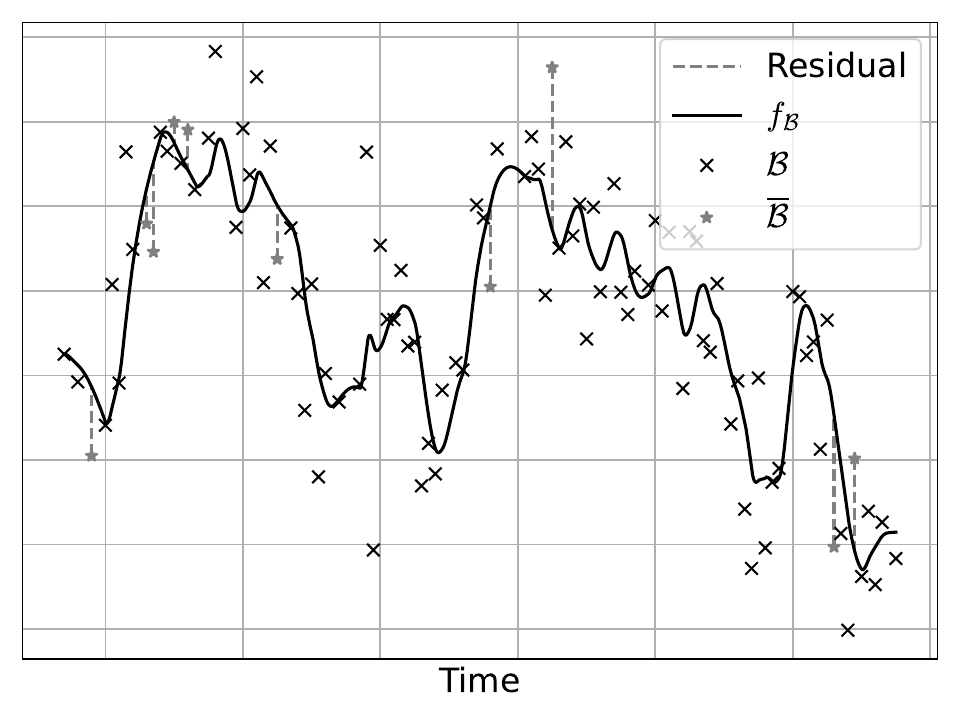}
    \caption{Illustration of the residuals used to estimate the MSE via bootstrapping. The test partition comes from the synthetic data and has been reconstructed with a Myopic(3,1) policy with $\eta=1$ and $\rho=2$. $\mathcal{B}$ contains $90\%$ of the test partition and $\overline{\mathcal{B}}$ the remaining $10\%$.}
    \label{fig:bootstrap}
\end{figure}

When the data distribution, or in our case, the underlying (assumed) smooth process, is unknown, we cannot follow the standard MSE and MAE computation procedure because the original function $\psi\in\mathcal{W}_\rho$ is also unknown.
Instead, we only have access to a certain dataset of test samples, e.g., $\mathcal{D} = \left\{\left(x_t,\psi(x_t)\right)\right\}^T_{t=1}$.
Following a bootstrap-inspired method \cite{efron1994introduction}, we choose a subset $\mathcal{B}\subseteq \mathcal{D}$ for the signal reconstruction and use the complementary set $\overline{\mathcal{B}}$, i.e., $\mathcal{B}\cup\overline{\mathcal{B}} = \mathcal{D}$ and $\mathcal{B}\cap\overline{\mathcal{B}} = \emptyset$ to estimate the MSE and MAE performance metrics.
Mathematically, this can be expressed as:
\begin{subequations}
\begin{align}
    \text{MSE}\left( f_\mathcal{B} \right) = \frac{1}{|\overline{\mathcal{B}}|} \sum_{i\in\overline{\mathcal{B}}} \left( f_\mathcal{B}(x_i) - \psi(x_i) \right)^2 , \\
    \text{MAE}\left( f_\mathcal{B} \right) = \frac{1}{|\overline{\mathcal{B}}|} \sum_{i\in\overline{\mathcal{B}}} \left| f_\mathcal{B}(x_i) - \psi(x_i) \right| ,
\end{align}
\end{subequations}
where $f_\mathcal{B}$ is the signal estimate constructed from the test data subset $\mathcal{B}$.
This procedure is illustrated in Fig. \ref{fig:bootstrap}.

Notice that it is important to partition the test data because any test data sample used for the signal reconstruction cannot be used to compute the performance metrics.
Otherwise, this results in data leakage.
On the other hand, due to the lack of data samples, the performance metrics estimated in this way, may not be as accurate as if we had larger test sets, or more specifically, large test sets with higher temporal resolution.
Thus, to reduce the variance of the MSE and MAE estimators, we repeat the procedure for several randomly chosen partitions $\mathcal{B}, \Bar{\mathcal{B}}$ with replacement (i.e. they may repeat) and average the result.
Particularly, we perform $10$ repetitions.

    % BIBLIOGRAPHY
    \bibliography{references}

\end{document}